\makeatletter
\newcommand{\dontusepackage}[2][]{%
  \@namedef{ver@#2.sty}{9999/12/31}%
  \@namedef{opt@#2.sty}{#1}}
\makeatother
\dontusepackage{subfigure}
\documentclass[tablecaption=bottom,wcp]{jmlr}
\usepackage{amssymb,amsmath,amsbsy,bm}
\usepackage{fixltx2e} % provides \textsubscript
\IfFileExists{upquote.sty}{\usepackage{upquote}}{}
\usepackage[T1]{fontenc}
\usepackage[utf8]{inputenc}
\IfFileExists{microtype.sty}{\usepackage{microtype}}{}
\usepackage{natbib}
\usepackage{listings}
\usepackage{longtable,booktabs}
\usepackage{graphicx}
\makeatletter
\def\ScaleIfNeeded{%
  \ifdim\Gin@nat@width>\linewidth
    \linewidth
  \else
    \Gin@nat@width
  \fi
}
\makeatother
\let\Oldincludegraphics\includegraphics
{
 \catcode`\@=11\relax
 \gdef\includegraphics{\@ifnextchar[{\Oldincludegraphics}{\Oldincludegraphics[width=\ScaleIfNeeded]}}
}
\usepackage{caption}
\usepackage{float}
\usepackage{subfig}
\jmlrproceedings{AABI 2020}{3rd Symposium on Advances in Approximate Bayesian Inference, 2020}

\title[Preconditioned training of normalizing flows]{Preconditioned training of normalizing flows for variational inference
in inverse problems}

\author{
\Name{Ali Siahkoohi} \Email{alisk@gatech.edu} \\
  \addr{Georgia Institute of Technology}
  \AND
\Name{Gabrio Rizzuti}\nametag{\thanks{Now at Utrecht University.}} \Email{g.rizzuti@umcutrecht.nl} \\
  \addr{Georgia Institute of Technology}
  \AND
\Name{Mathias Louboutin} \Email{mlouboutin3@gatech.edu} \\
  \addr{Georgia Institute of Technology}
  \AND
\Name{Philipp A. Witte} \Email{pwitte@microsoft.com} \\
  \addr{Microsoft}
  \AND
\Name{Felix J. Herrmann} \Email{felix.herrmann@gatech.edu} \\
  \addr{Georgia Institute of Technology}
  \AND
}

\date{}

\begin{document}
\maketitle

\begin{abstract}
Obtaining samples from the posterior distribution of inverse problems
with expensive forward operators is challenging especially when the
unknowns involve the strongly heterogeneous Earth. To meet these
challenges, we propose a preconditioning scheme involving a conditional
normalizing flow (NF) capable of sampling from a low-fidelity posterior
distribution directly. This conditional NF is used to speed up the
training of the high-fidelity objective involving minimization of the
Kullback-Leibler divergence between the predicted and the desired
high-fidelity posterior density for indirect measurements at hand. To
minimize costs associated with the forward operator, we initialize the
high-fidelity NF with the weights of the pretrained low-fidelity NF,
which is trained beforehand on available model and data pairs. Our
numerical experiments, including a 2D toy and a seismic compressed
sensing example, demonstrate that thanks to the preconditioning
considerable speed-ups are achievable compared to training NFs from
scratch.
\end{abstract}

\section{Introduction}\label{introduction}

Our aim is to perform approximate Bayesian inference for inverse
problems characterized by computationally expensive forward operators,
$F: \mathcal{X} \rightarrow \mathcal{Y}$, with a data likelihood,
$\pi_{\text{like}} (\boldsymbol{y} \mid \boldsymbol{x})$:
\begin{equation}
\boldsymbol{y} = F (\boldsymbol{x}) + \boldsymbol{\epsilon},
\label{fwd-op}
\end{equation}
 where $\boldsymbol{x} \in \mathcal{X}$ is the unknown model,
$\boldsymbol{y} \in \mathcal{Y}$ the observed data, and
$\boldsymbol{\epsilon} \sim \mathrm{N}(\boldsymbol{0}, \sigma^2 \boldsymbol{I})$
the measurement noise. Given a prior density,
$\pi_{\text{prior}}(\boldsymbol{x})$, variational inference
\citep[VI,][]{jordan1999introduction} based on normalizing flows
\citep[NFs,][]{rezende2015variational} can be used where the
Kullback-Leibler (KL) divergence is minimized between the predicted and
the target---i.e., \emph{high-fidelity}, posterior density
$\pi_{\text{post}} (\boldsymbol{x} \mid \boldsymbol{y} )$
\citep{liu2016stein, kruse2019hint, rizzuti2020SEGpub, siahkoohi2020TRfuqf, sun2020deep}:
\begin{equation}
\min_{\theta}\, \mathbb{E}\,_{\boldsymbol{z} \sim
    \pi_{z}(\boldsymbol{z})}
    \bigg [ \frac{1}{2\sigma^2}  \left \| F \big(T_{\theta}
    (\boldsymbol{z}) \big) -
    \boldsymbol{y} \right \|_2^2  -\log \pi_{\text{prior}}
    \big (T_{\theta} (\boldsymbol{z}) \big ) -\log  \Big | \det
    \nabla_{z} T_{\theta} (\boldsymbol{z}) \Big | \bigg ].
\label{hint2-obj}
\end{equation}
 In the above expression, $T_{\theta} : \mathcal{Z}_x \to \mathcal{X}$
denotes a NF with parameters $\boldsymbol{\theta}$ and a Gaussian latent
variable $\boldsymbol{z} \in \mathcal{Z}_x$. The above objective
consists of the data likelihood term, regularization on the output of
the NF, and a log-determinant term that is related to the entropy of the
NF output. The last term is necessarily to prevent the output of the NF
from collapsing on the maximum a posteriori estimate. For details
regarding the derivation of the objective in Equation (\ref{hint2-obj}),
we refer to Appendix A. During training, we replace the expectation by
Monte-Carlo averaging using mini-batches of $\boldsymbol{z}$. After
training, samples from the approximated posterior,
$\pi_{\theta}(\boldsymbol{x} \mid \boldsymbol{y}) \approx \pi_{\text{post}} (\boldsymbol{x} \mid \boldsymbol{y})$,
can be drawn by evaluating $T_{\theta}(\boldsymbol{z})$ for
$\boldsymbol{z} \sim \pi_{z}(\boldsymbol{z})$ \citep{kruse2019hint}. It
is important to note that Equation (\ref{hint2-obj}) trains a NF
specific to the observed data $\boldsymbol{y}$. While the above VI
formulation in principle allows us to train a NF to generate samples
from the posterior given a single observation $\boldsymbol{y}$, this
variational estimate requires access to a prior density, and the
training calls for repeated evaluations of the forward operator, $F$, as
well as the adjoint of its Jacobian, $\nabla {F}^\top$. As in
multi-fidelity Markov chain Monte Carlo (MCMC) sampling
\citep{marzouk2018multifidelity}, the costs associated with the forward
operator may become prohibitive even though VI-based methods are known
to have computational advantages over MCMC \citep{blei2017variational}.

\begin{figure}
\centering
\subfloat[\label{shallow-1}]{\includegraphics[width=0.230\hsize]{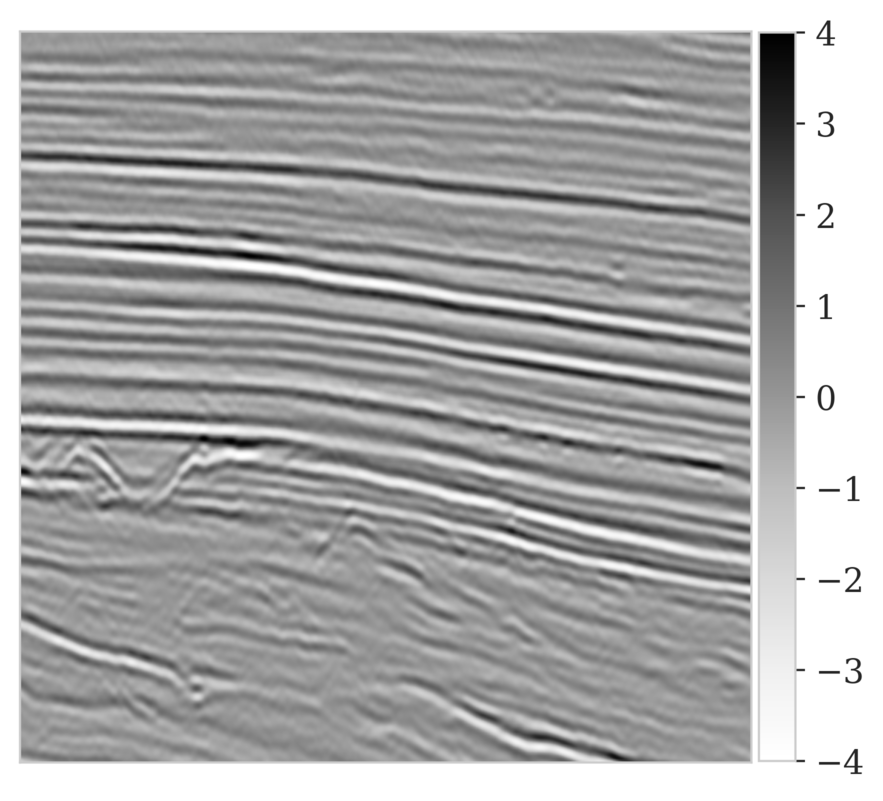}}
\subfloat[\label{shallow-2}]{\includegraphics[width=0.230\hsize]{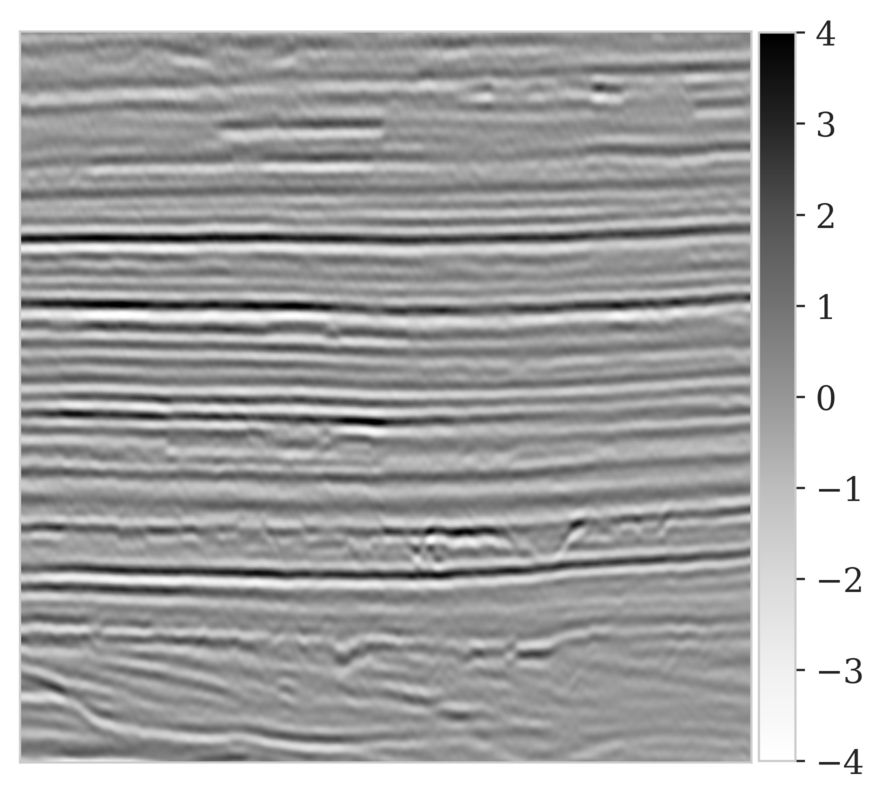}}
\subfloat[\label{deep-1}]{\includegraphics[width=0.230\hsize]{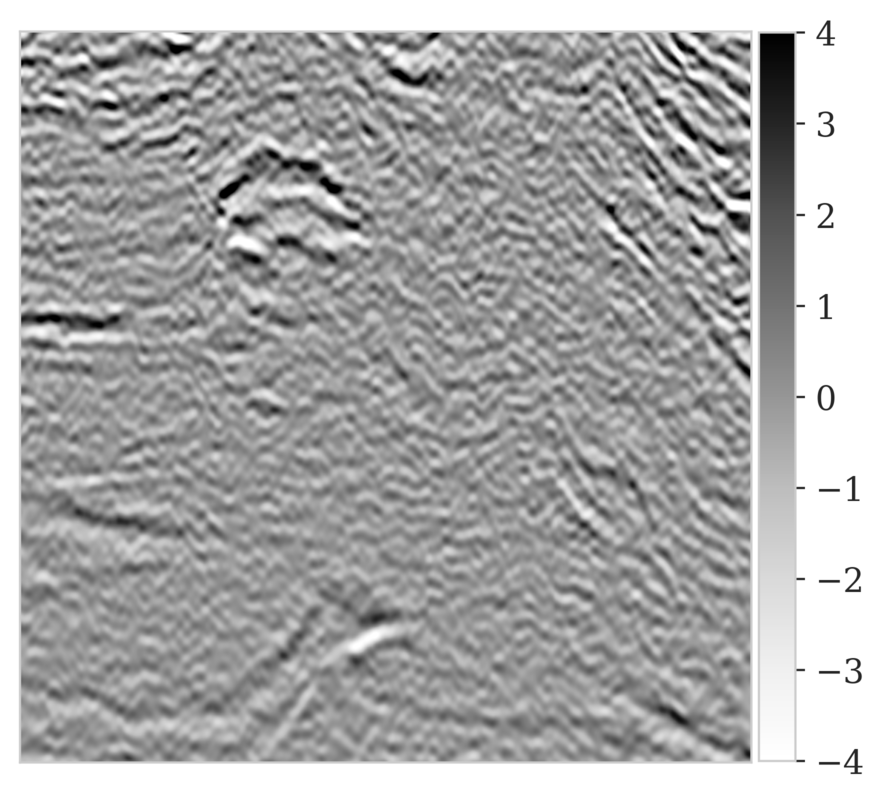}}
\subfloat[\label{deep-2}]{\includegraphics[width=0.230\hsize]{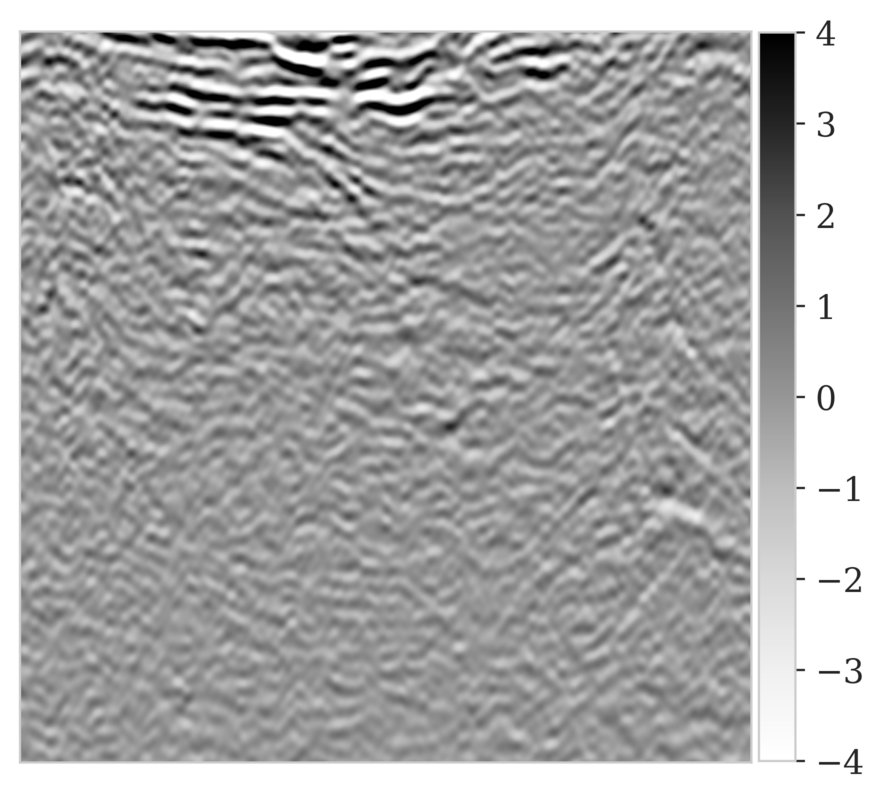}}
\caption{Subsurface images estimated from a real seismic survey,
indicating strong heterogeneity between (a), (b) shallow and (c), (d)
deep parts of the same survey area.}\label{high-vs-low-fidelity}
\end{figure}

Aside from the above computational considerations, reliance on having
access to a prior may be problematic especially when dealing with images
of the Earth's subsurface, which are the result of complex geological
processes that do not lend themselves to be easily captured by
hand-crafted priors. Under these circumstances, data-driven priors---or
even better data-driven posteriors obtained by training over model and
data pairs sampled from the joint distribution,
$\widehat{\pi}_{y, x} (\boldsymbol{y}, \boldsymbol{x})$---are
preferable. More specifically, we follow \citet{kruse2019hint},
\citet{kovachki2020conditional}, and \citet{baptista2020adaptive}, and
formulate the objective function in terms of a block-triangular
conditional NF,
$G_{\phi} : \mathcal{Y} \times \mathcal{X} \to \mathcal{Z}_y \times \mathcal{Z}_x$,
with latent space $\mathcal{Z}_y \times \mathcal{Z}_x$:
\begin{equation}
\begin{aligned}
& \min_{\phi}\, \mathbb{E}\,_{\boldsymbol{y}, \boldsymbol{x}
    \sim \widehat{\pi}_{y, x} (\boldsymbol{y}, \boldsymbol{x})}\,
    \left [ \frac{1}{2} \left \| G_{\phi} (\boldsymbol{y}, \boldsymbol{x}) \right \|^2 - \log \Big|\det \nabla_{y, x}\, G_{\phi} (\boldsymbol{y}, \boldsymbol{x}) \Big | \right ], \\
& \text{where} \quad G_{\phi}(\boldsymbol{y}, \boldsymbol{x})
    =\begin{bmatrix}  G_{\phi_y} (\boldsymbol{y}) \\G_{\phi_x}
    (\boldsymbol{y}, \boldsymbol{x}) \end{bmatrix}, \
    \boldsymbol{\phi} = \left \{ \boldsymbol{\phi}_y ,
    \boldsymbol{\phi}_x \right \}.
\end{aligned}
\label{hint3-obj}
\end{equation}
 Thanks to the block-triangular structure of $G_{\phi}$, samples of the
approximated posterior,
$\pi_{\phi}(\boldsymbol{x} \mid \boldsymbol{y}) \approx \pi_{\text{post}} (\boldsymbol{x} \mid \boldsymbol{y})$
can be drawn by evaluating
$G_{\phi_x}^{-1} (G_{\phi_y} (\boldsymbol{y}), \boldsymbol{z})$ for
$\boldsymbol{z} \sim \pi_{z}(\boldsymbol{z})$
\citep{marzouk2016sampling}. Unlike the objective in Equation
(\ref{hint2-obj}), training $G_{\phi}$ does not involve multiple
evaluations of $F$ and $\nabla F^\top$, nor does it require specifying a
prior density. However, its success during inference heavily relies on
having access to training pairs from the joint distribution,
$\boldsymbol{y}, \boldsymbol{x}\sim \widehat{\pi}_{y, x} (\boldsymbol{y}, \boldsymbol{x})$.
Unfortunately, unlike medical imaging, where data is abundant and
variability among patients is relatively limited, samples from the joint
distribution are unavailable in geophysical applications. Attempts have
been made to address this lack of training pairs including the
generation of simplified artificial geological models
\citep{mosser2018stochastic}, but these approaches cannot capture the
true heterogeneity exhibited by the Earth's subsurface. This is
illustrated in Figure~\ref{high-vs-low-fidelity}, which shows several
true seismic image patches drawn from the
\href{https://wiki.seg.org/wiki/Parihaka-3D}{Parihaka} dataset. Even
though samples are drawn from a single data set, they illustrate
significant differences between shallow (Figures~\ref{shallow-1}
and~\ref{shallow-2}) and deeper (Figures~\ref{deep-1} and~\ref{deep-2})
sections.

To meet the challenges of computational cost, heterogeneity and lack of
access to training pairs, we propose a preconditioning scheme where the
two described VI methods are combined to:

\begin{enumerate}
\def\labelenumi{\arabic{enumi}.}
\item
  take maximum advantage of available samples from the joint
  distribution $\widehat{\pi}_{y, x} (\boldsymbol{y}, \boldsymbol{x})$,
  to pretrain $G_{\phi}$ by minimizing Equation (\ref{hint3-obj}). We
  only incur these costs once, by training this NF beforehand. As these
  samples typically come from a different (neighboring) region, they are
  considered as low-fidelity;
\item
  exploit the invertibility of $G_{\phi_x} (\boldsymbol{y}, \,\cdot\,)$,
  which gives us access to a low-fidelity posterior density,
  $\pi_{\phi}(\boldsymbol{x} \mid \boldsymbol{y})$. For a given
  $\boldsymbol{y}$, this trained (conditional) prior can be used in
  Equation (\ref{hint2-obj});
\item
  initialize $T_{\theta}$ with weights from the pretrained
  $G_{\phi_x}^{-1}$. This initialization can be considered as an
  instance of transfer learning \citep{yosinski2014transferable}, and we
  expect a considerable speed-up when solving
  Equation~\eqref{hint2-obj}. This is important since it involves
  inverting $F$, which is computationally expensive.
\end{enumerate}

\section{Related work}\label{related-work}

In the context of variational inference for inverse problems with
expensive forward operators, \citet{herrmann2019NIPSliwcuc} train a
generative model to sample from the posterior distribution, given
indirect measurements of the unknown model. This approach is based on an
Expectation Maximization technique, which infers the latent
representation directly instead of using an inference encoding model.
While that approach allows for inclusion of hand-crafted priors,
capturing the posterior is not fully developed. Like
\citet{kovachki2020conditional}, we also use a block-triangular map
between the joint model and data distribution and their respective
latent spaces to train a network to generate samples from the
conditional posterior. By imposing an additional monotonicity
constraint, these authors train a generative adversarial network
\citep[GAN,][]{Goodfellow2014} to directly sample from the posterior
distribution. To allow for scalability to large scale problems, we work
with NFs instead, because they allow for more memory efficient training
\citep{vandeLeemput2019MemCNN, putzky2019invert, peters2020fullysensing, peters2020fully}.
Our contribution essentially corresponds to a reformulation of
\citet{parno2018transport} and \citet{marzouk2018multifidelity}. In that
work, transport-based maps are used as non-Gaussian proposal
distributions during MCMC sampling. As part of the MCMC, this transport
map is then tuned to match the target density, which improves the
efficiency of the sampling. \citet{marzouk2018multifidelity} extend this
approach by proposing a preconditioned MCMC sampling technique where a
transport-map trained to sample from a low-fidelity posterior
distribution is used as a preconditioner. This idea of multi-fidelity
preconditioned MCMC inspired our work where we setup a VI objective
instead. We argue that this formulation can be faster and may be easier
to scale to large-scale Bayesian inference problems
\citep{blei2017variational}.

Finally, there is a conceptional connection between our work and
previous contributions on amortized variational inference
\citep{gershman2014amortized}, including an iterative refinement step
\citep{hjelm2016iterative, pmlrv84krishnan18a, pmlrv80kim18e, marino2018iterative}.
Although similar in spirit, our approach is different from these
attempts because we adapt the weights of our conditional generative
model to account for the inference errors instead of correcting the
inaccurate latent representation of the out-of-distribution data.

\section{Multi-fidelity preconditioning
scheme}\label{multi-fidelity-preconditioning-scheme}

For an observation $\boldsymbol{y}$, we define a NF
$T_{\phi_x} : \mathcal{Z}_x \to \mathcal{X}$ as
\begin{equation}
T_{\phi_x}(\boldsymbol{z}):= G_{\phi_x}^{-1} (G_{\phi_y} (\boldsymbol{y}), \boldsymbol{z}),
\label{low-fidelity-NF}
\end{equation}
 where we obtain
$\boldsymbol{\phi} = \left \{ \boldsymbol{\phi}_y , \boldsymbol{\phi}_x \right \}$
by training $G_{\phi}$ through minimizing the objective function in
Equation (\ref{hint3-obj}). To train $G_{\phi}$, we use available
low-fidelity training pairs
$\boldsymbol{y}, \boldsymbol{x} \sim \widehat{\pi}_{y, x} (\boldsymbol{y}, \boldsymbol{x})$.
We perform this training phase beforehand, similar to the pretraining
phase during transfer learning \citep{yosinski2014transferable}. Thanks
to the invertibility of $G_{\phi}$, it provides an expression for the
posterior. We refer to this posterior as low-fidelity because the
network is trained with often scarce and out-of-distribution training
pairs. Because the Earth's heterogeneity does not lend itself to be
easily captured by hand-crafted priors, we argue that this NF can still
serve as a (conditional) prior in Equation (\ref{hint2-obj}):
\begin{equation}
\pi_{\text{prior}} (\boldsymbol{x}) := \pi_{\boldsymbol{z}} \big(G_{\phi_x} (\boldsymbol{y}, \boldsymbol{x}) \big )\,
\Big | \det \nabla_{x} G_{\phi_x} (\boldsymbol{y}, \boldsymbol{x}) \Big |.
\label{low-fidelity-post}
\end{equation}
 To train the high-fidelity NF given observed data $\boldsymbol{y}$, we
minimize the KL divergence between the predicted and the high-fidelity
posterior density,
$\pi_{\text{post}} (\boldsymbol{x} \mid \boldsymbol{y} )$
\citep{liu2016stein, kruse2019hint}
\begin{equation}
\min_{\phi_x}\, \mathbb{E}_{\boldsymbol{z} \sim
    \pi_{z}(\boldsymbol{z})}
    \bigg [ \frac{1}{2\sigma^2}  \left \| F \big(T_{\phi_x}
    (\boldsymbol{z}) \big) -
    \boldsymbol{y} \right \|_2^2  -\log \pi_{\text{prior}}
    \big (T_{\phi_x} (\boldsymbol{z}) \big ) -\log  \Big | \det
    \nabla_{z} T_{\phi_x} (\boldsymbol{z}) \Big | \bigg ],
\label{hint2-obj-finetune}
\end{equation}
 where the prior density of Equation (\ref{low-fidelity-post}) is used.
Notice that this minimization problem differs from the one stated in
Equation (\ref{hint2-obj}). Here, the optimization involves
``fine-tuning'' the low-fidelity network parameters $\phi_x$ introduced
in Equation (\ref{hint3-obj}). Moreover, this low-fidelity network is
also used as a prior. While other choices exist for the latter---e.g.,
it could be replaced or combined with a hand-crafted prior in the form
of constraints \citep{peters2018pmf} or by a separately trained
data-driven prior \citep{mosser2018stochastic}, using the low-fidelity
posterior as a prior (cf.~Equation (\ref{low-fidelity-post})) has
certain distinct advantages. First, it removes the need for training a
separate data-driven prior model. Second, use of the low-fidelity
posterior may be more informative \citep{yang2018conditional} than its
unconditional counterpart because it is conditioned by the observed data
$\boldsymbol{y}$. In addition, our multi-fidelity approach has strong
connections with online variational Bayes \citep{zeno2018task} where
data arrives sequentially and previous posterior approximates are used
as priors for subsequent approximations.

In summary, the problem in Equation~\eqref{hint2-obj-finetune} can be
interpreted as an instance of transfer learning
\citep{yosinski2014transferable} for conditional NFs. This formulation
is particularly useful for inverse problems with expensive forward
operators, where access to high fidelity training samples, i.e.~samples
from the target distribution, is limited. In the next section, we
present two numerical experiments designed to show the speed-up and
accuracy gained with our proposed multi-fidelity formulation.

\section{Numerical experiments}\label{numerical-experiments}

We present two synthetic examples aimed at verifying the anticipated
speed-up and increase in accuracy of the predicted posterior density via
our multi-fidelity preconditioning scheme. The first example is a
two-dimensional problem where the posterior density can be accurately
and cheaply sampled via MCMC. The second example demonstrates the effect
of the preconditioning scheme in a seismic compressed sensing
\citep{candes2006stable, donoho2006compressed} problem. Details
regarding training hyperparameters and the NF architectures are included
in Appendix B. Code to reproduce our results are made available on
\href{https://github.com/slimgroup/Software.siahkoohi2021AABIpto}{GitHub}.
Our implementation relies on
\href{https://github.com/slimgroup/InvertibleNetworks.jl}{InvertibleNetworks.jl}
\citep{invnet}, a recently-developed memory-efficient framework for
training invertible networks in the Julia programming language.

\subsection{2D toy example}\label{d-toy-example}

\begin{figure}
\centering
\subfloat[\label{toy-model}]{\includegraphics[width=0.250\hsize]{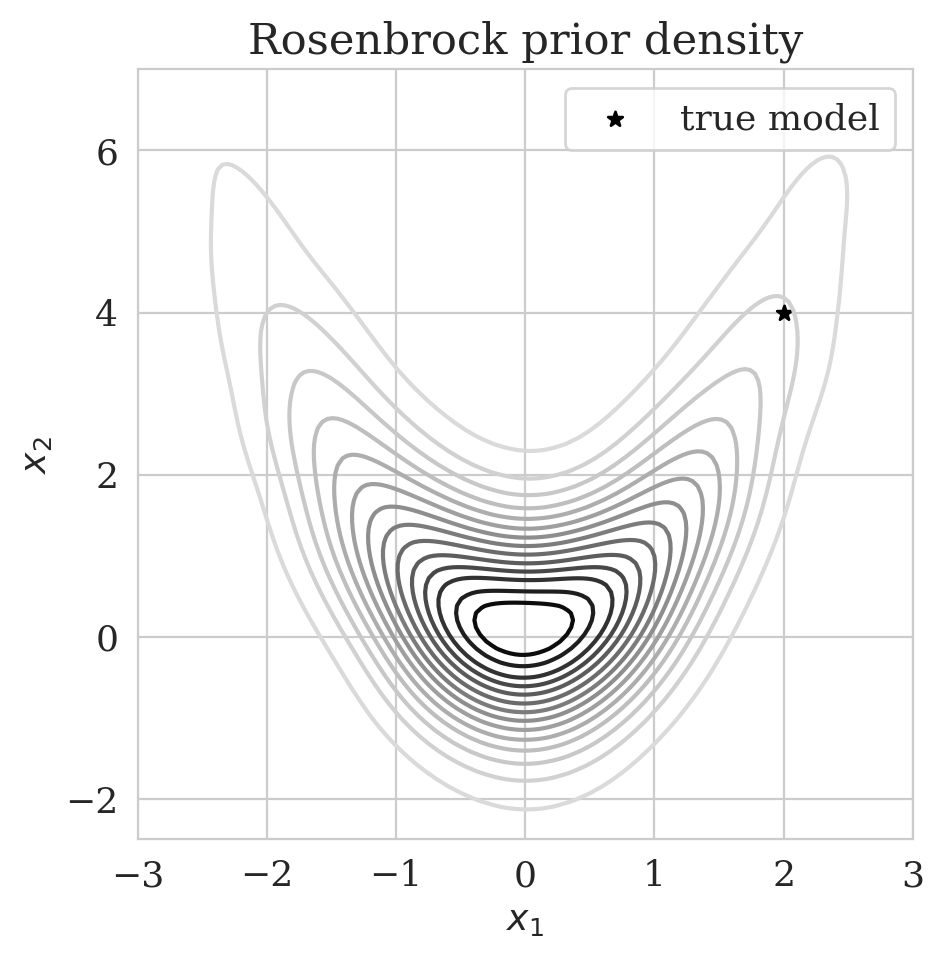}}
\subfloat[\label{toy-data}]{\includegraphics[width=0.250\hsize]{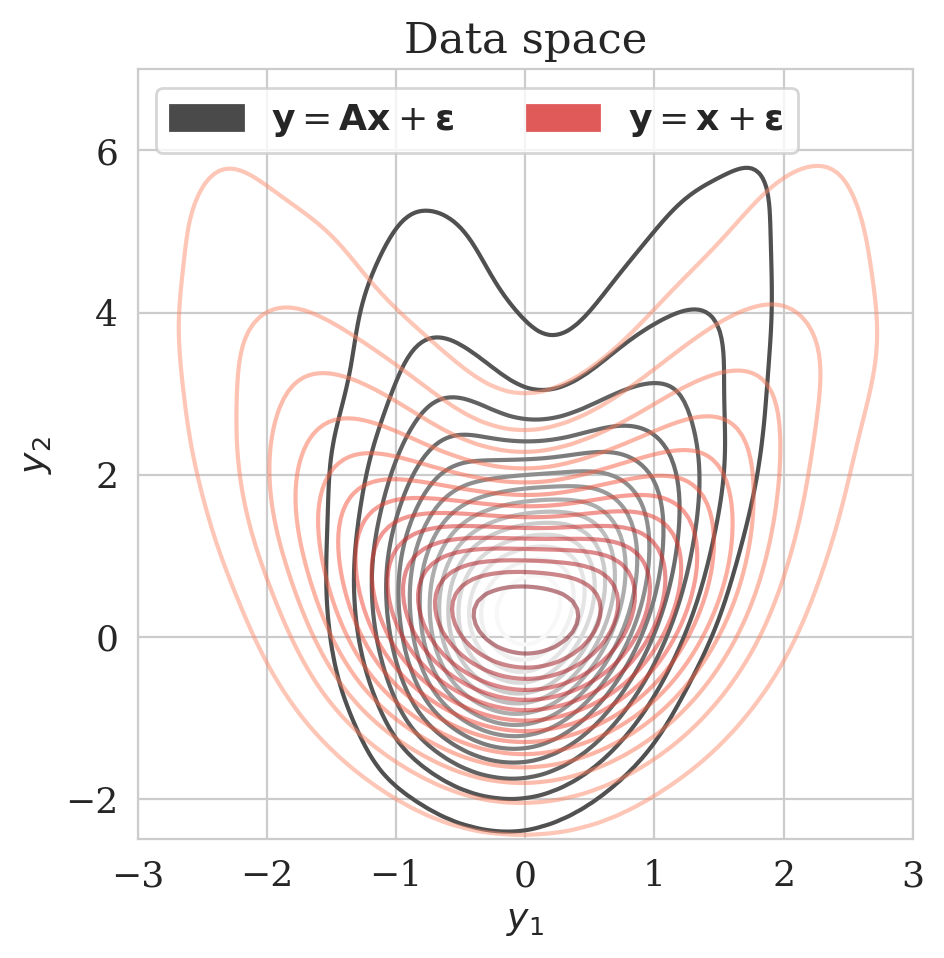}}
\subfloat[\label{prior-posterior-samples}]{\includegraphics[width=0.260\hsize]{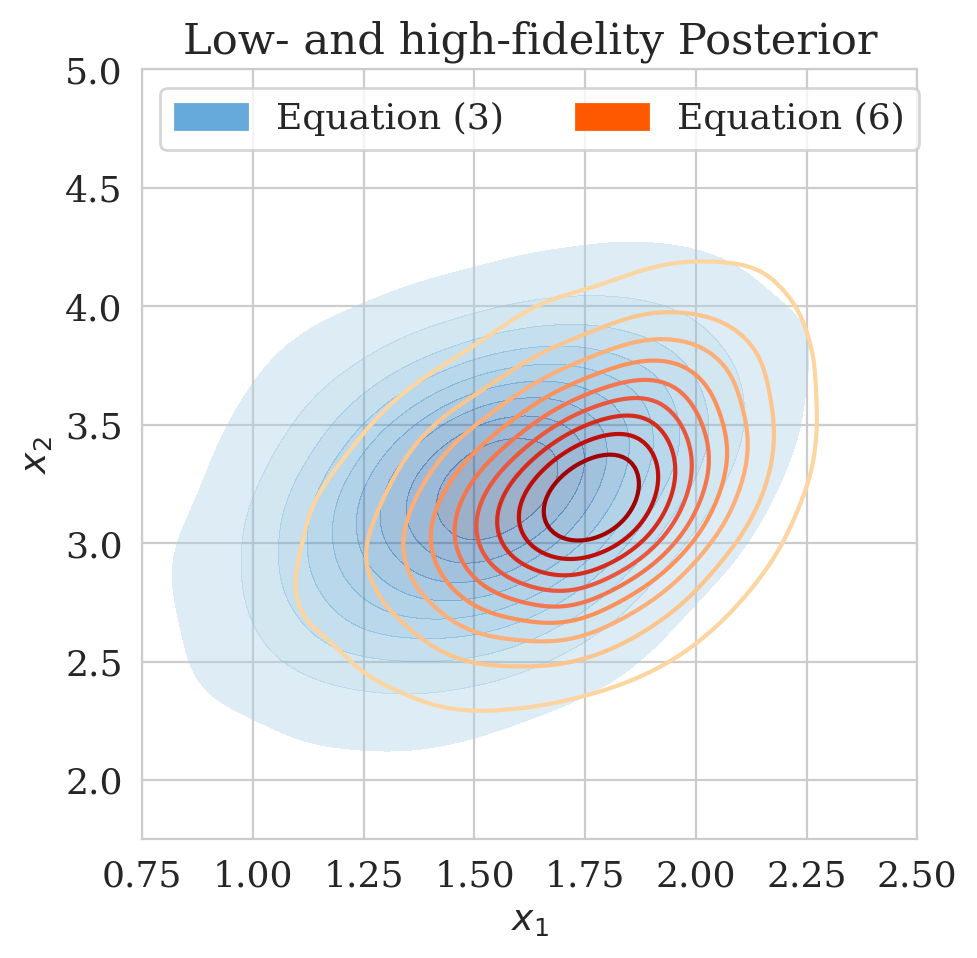}}
\\
\subfloat[\label{toy-post-scratch}]{\includegraphics[width=0.260\hsize]{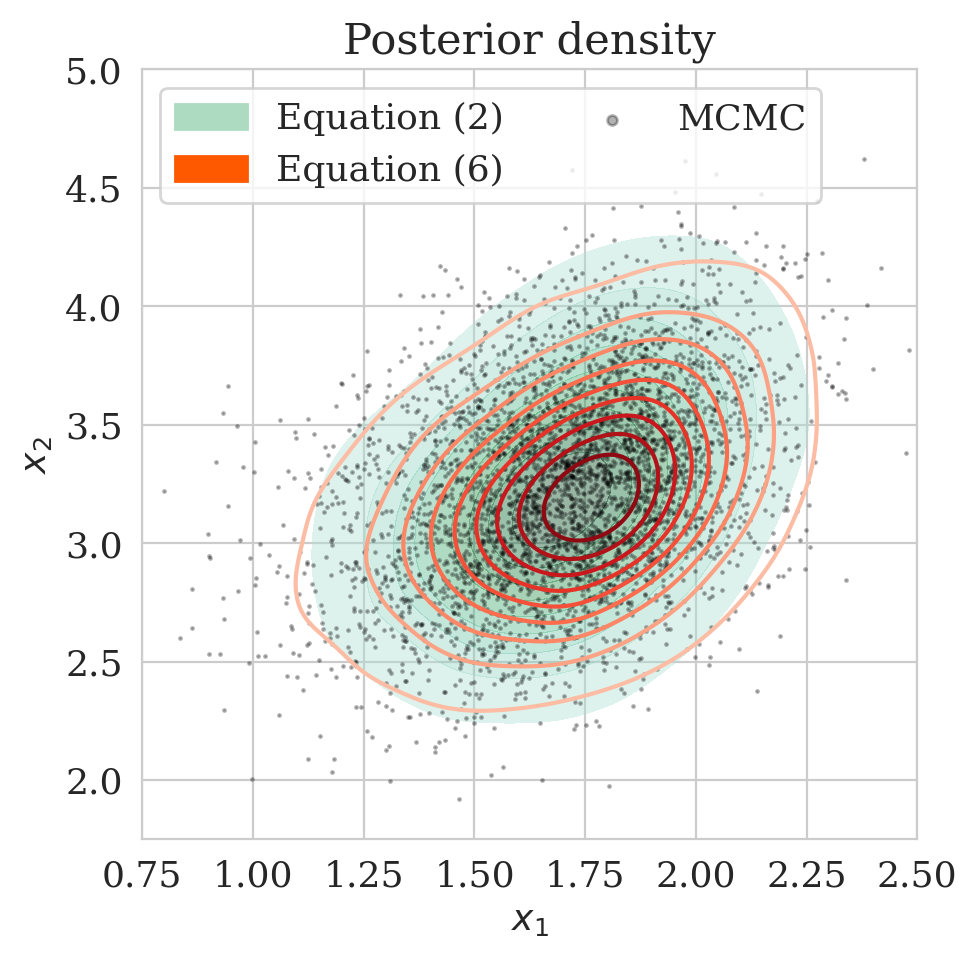}}
\subfloat[\label{toy-obj}]{\includegraphics[width=0.245\hsize]{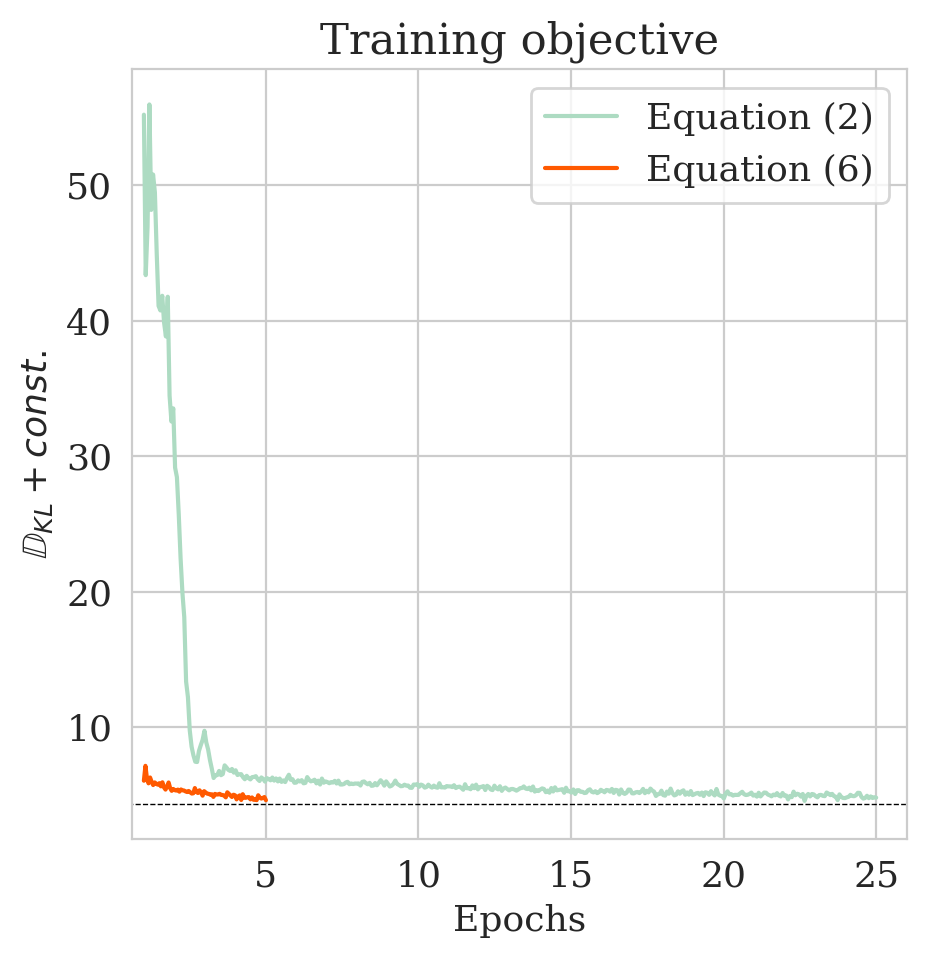}}
\caption{(a) Prior, (b) low- and high-fidelity data, (c) low- (blue) and
high-fidelity (orange) approximated posterior densities, and (d)
approximated posterior densities via MCMC (dark circles), and objectives
in Equations (\ref{hint2-obj}) in green and (\ref{hint2-obj-finetune})
in orange. (e) Objective value during training via Equations
(\ref{hint2-obj}) in green and (\ref{hint2-obj-finetune}) in
orange.}\label{rosenbrock-samples}
\end{figure}

To illustrate, the advantages of working with our multi-fidelity scheme,
we consider the 2D Rosenbrock distribution,
$\pi_{\text{prior}} (\boldsymbol{x}) \propto \exp \left( - \frac{1}{2} x_{1}^{2} - \left(x_{2}-x_{1}^{2} \right)^{2}\right)$,
plotted in Figure~\ref{toy-model}. High-fidelity data
$\boldsymbol{y} \in\mathbb{R}^{2}$ are generated via
$\boldsymbol{y} = \boldsymbol{A}\, \boldsymbol{x} + \boldsymbol{\epsilon}$,
where
$\boldsymbol{\epsilon} \sim \mathrm{N}(\boldsymbol{0}, 0.4^2 \boldsymbol{I})$
and $\boldsymbol{A} \in \mathbb{R}^{2 \times 2}$ is a forward operator.
To control the discrepancy between the low- and high-fidelity samples,
we set $\boldsymbol{A}$ equal to
$\bar{\boldsymbol{A}} \mathbin{/} \rho (\bar{\boldsymbol{A}})$, where
$\rho ( \cdot)$ is the spectral radius of
$\bar{\boldsymbol{A}} = \boldsymbol{\Gamma} + \gamma \boldsymbol{I}$,
$\boldsymbol{\Gamma} \in \mathbb{R}^{2 \times 2}$ has independent and
normally distributed entries, and $\gamma = 3$. By choosing smaller
values for $\gamma$, we make $\boldsymbol{A}$ more dissimilar to the
identity matrix, therefore increasing the discrepancy between the low-
and high-fidelity posterior.

Figure~\ref{toy-data} depicts the low- (purple) and high-fidelity (red)
data densities. The dark star represents the unknown model. Low-fidelity
data samples are generated with the identity as the forward operator.
During the pretraining phase conducted beforehand, we minimize the
objective function in Equation (\ref{hint3-obj}) for $25$ epochs.

The pretrained low-fidelity posterior is subsequently used to
precondition the minimization of (\ref{hint2-obj-finetune}) given
observed data $\boldsymbol{y}$. The resulting low- and high-fidelity
estimates for the posterior as plotted in
Figure~\ref{prior-posterior-samples}~differ significantly. In Figure
(\ref{toy-post-scratch}), the accuracy of the proposed method is
verified by comparing the approximated high-fidelity posterior density
(orange contours) with the approximation (in green) obtained by
minimizing the objective of Equation~\eqref{hint2-obj}. The overlap
between the orange contours and the green shaded area confirms
consistency between the two methods. To assess the accuracy of the
estimated densities themselves, we also include samples from the
posterior (dark circles) obtained via stochastic gradient Langevin
dynamics \citep{welling2011bayesian}, an MCMC sampling technique. As
expected, the estimated posterior densities with and without the
preconditioning scheme are in agreement with the MCMC samples.

Finally, to illustrate the performance our multi-fidelity scheme, we
consider the convergence plot in Figure~\ref{toy-obj} where the
objective values of Equations (\ref{hint2-obj}) and
(\ref{hint2-obj-finetune}) are compared. As explained in Appendix A, the
values of the objective functions correspond to the KL divergence (plus
a constant) between the posterior given by Equation (\ref{hint2-obj})
and the posterior distribution obtained by our multi-fidelity approach
(Equation (\ref{hint2-obj-finetune})). As expected, the multi-fidelity
objective converges much faster because of the ``warm start''. In
addition, the updates of $T_{\phi_x}$ via Equation
(\ref{hint2-obj-finetune}) succeed in bringing down the KL divergence
within only five epochs (see orange curve), whereas it takes $25$ epochs
via the objective in Equation~\eqref{hint2-obj} to reach approximately
the same KL divergence. This pattern holds for smaller values of
$\gamma$ too as indicated in Table~\ref{KL-divs}. According to
Table~\ref{KL-divs}, the improvements by our multi-fidelity method
become more pronounced if we decrease the $\gamma$. This behavior is to
be expected since the samples used for pretraining are more and more out
of distribution in that case. We refer to Appendix C for additional
figures for different values of $\gamma$.

\begin{table}
\centering
\begin{tabular}{cccccc}
\toprule\addlinespace
$\gamma$ & Low-fidelity & Without preconditioning & With
preconditioning\tabularnewline
\midrule
$3$ & $6.13$ & $4.88$ & $4.66$\tabularnewline
$2$ & $8.43$ & $5.60$ & $5.26$\tabularnewline
$1$ & $8.84$ & $6.26$ & $6.51$\tabularnewline
$0$ & $14.73$ & $8.41$ & $8.45$\tabularnewline
\bottomrule
\end{tabular}
\caption{The KL divergence (plus some constant, see Appendix A) between
different estimated posterior densities and high-fidelity posterior
distribution for different values of $\gamma$. Second column corresponds
to the low-fidelity posterior estimate obtained via
Equation~\eqref{hint3-obj}, third column relates to the posterior
estimate via Equation~\eqref{hint2-obj}, and finally, the last column
shows the same quantity for the posterior estimate via the
multi-fidelity preconditioned scheme.}\label{KL-divs}
\end{table}

\subsection{Seismic compressed sensing
example}\label{seismic-compressed-sensing-example}

This experiment is designed to show challenges with geophysical inverse
problems due to the Earth's strong heterogeneity. We consider the
inversion of noisy indirect measurements of image patches
$\boldsymbol{x} \in \mathbb{R}^{256 \times 256}$ sampled from deeper
parts of the \href{https://wiki.seg.org/wiki/Parihaka-3D}{Parihaka}
seismic dataset. The observed measurements are given by
$\boldsymbol{y} = \boldsymbol{A} \boldsymbol{x} + \boldsymbol{\epsilon}$
where
$\boldsymbol{\epsilon} \sim \mathrm{N}(\boldsymbol{0}, 0.2^2 \boldsymbol{I})$.
For simplicity, we chose
$\boldsymbol{A} = {\boldsymbol{M}}^T {\boldsymbol{M}}$ with
${\boldsymbol{M}}$ a compressing sensing matrix with $66.66\,$\%
subsampling. The measurement vector $\boldsymbol{y}$ corresponds to a
pseudo-recovered model contaminated with noise.

\begin{figure}
\centering
\subfloat[\label{denoising-true}]{\includegraphics[width=0.250\hsize]{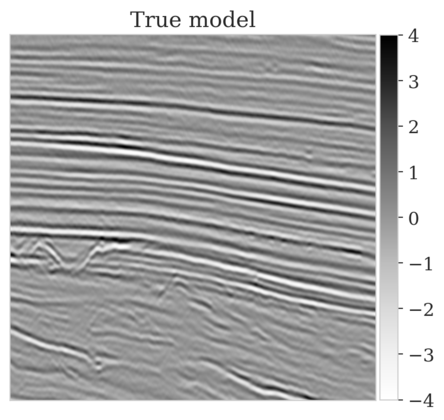}}
\subfloat[\label{denoising-obs}]{\includegraphics[width=0.250\hsize]{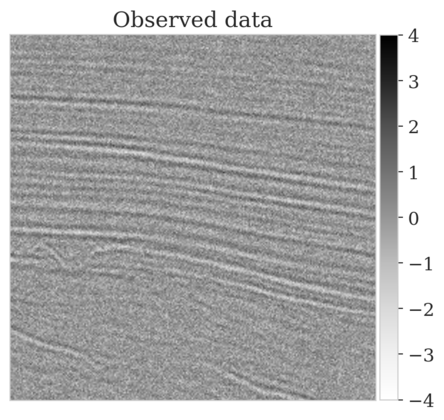}}
\subfloat[\label{denoising-cm}]{\includegraphics[width=0.250\hsize]{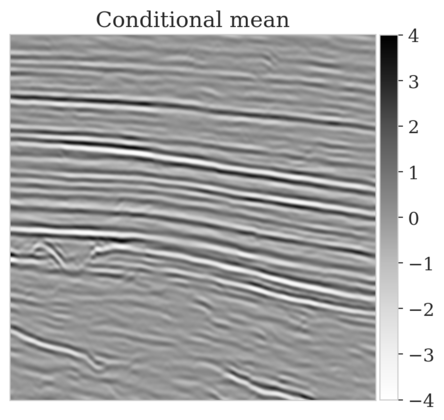}}
\subfloat[\label{denoising-std}]{\includegraphics[width=0.250\hsize]{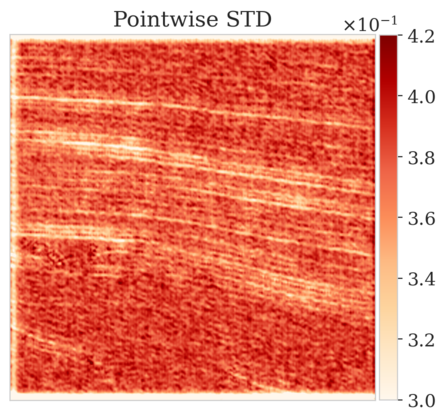}}
\\
\subfloat[\label{denoising-WS-true}]{\includegraphics[width=0.250\hsize]{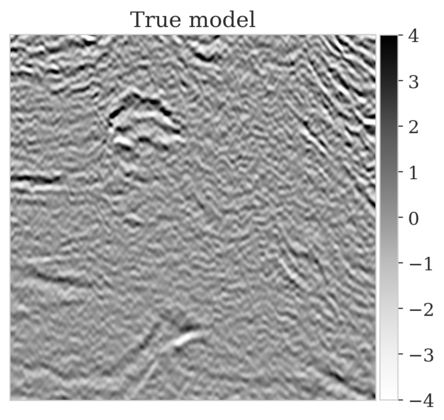}}
\subfloat[\label{denoising-WS-obs}]{\includegraphics[width=0.250\hsize]{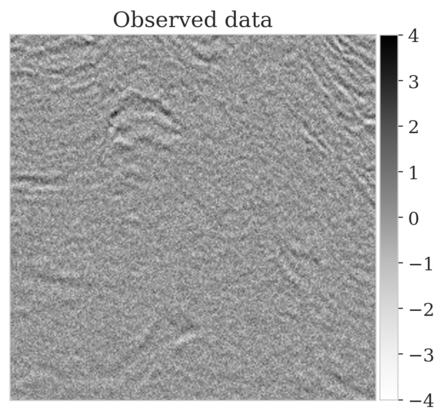}}
\subfloat[\label{denoising-cm-pretrained}]{\includegraphics[width=0.250\hsize]{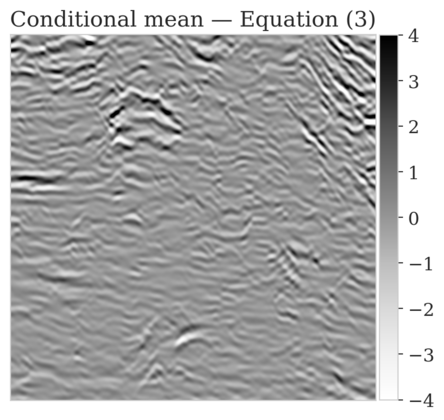}}
\subfloat[\label{denoising-cm-ws}]{\includegraphics[width=0.250\hsize]{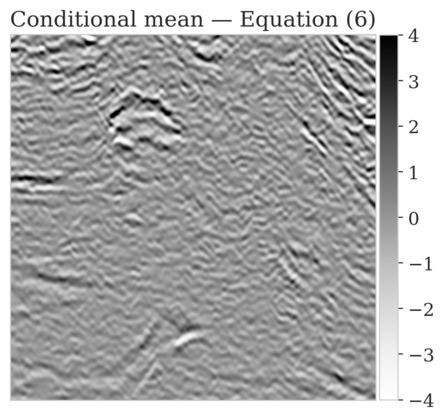}}
\\
\subfloat[\label{denoising-cm-err}]{\includegraphics[width=0.250\hsize]{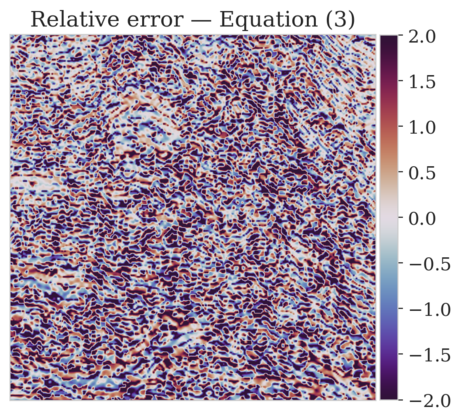}}
\subfloat[\label{denoising-cm-ws-err}]{\includegraphics[width=0.250\hsize]{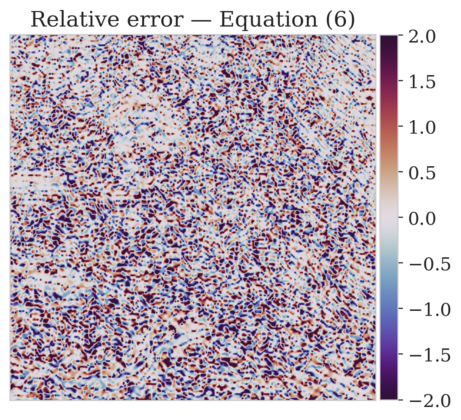}}
\subfloat[\label{denoising-std-ws-pretrained}]{\includegraphics[width=0.250\hsize]{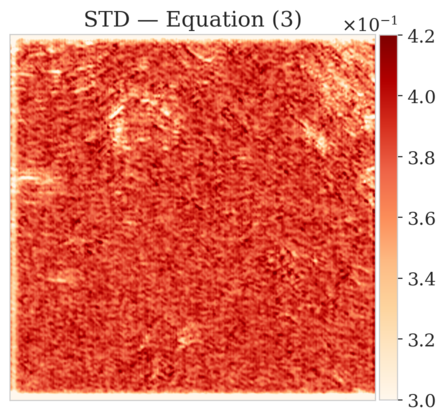}}
\subfloat[\label{denoising-std-ws}]{\includegraphics[width=0.250\hsize]{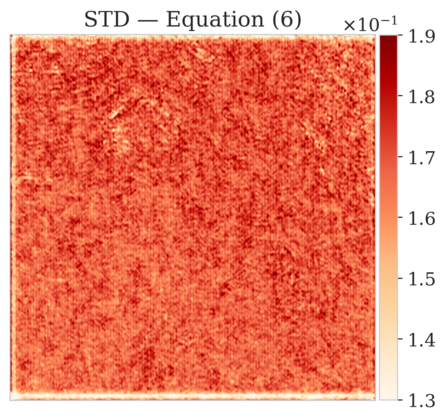}}
\caption{Seismic compressed sensing. First row indicates the performance
of the pretrained network on data not used during pretraining. Second
row is compares the pretraining-based recovery with the accelerated
scheme result. Last row compares the recovery errors and pointwise STDs
for the two recoveries.}\label{denoising-samples}
\end{figure}

To mimic a realistic situation in practice, we change the likelihood
distribution by reducing the standard deviation of the noise to $0.01$
in combination with using image patches sampled from the shallow part of
the Parihaka dataset. As we have seen in
Figure~\ref{high-vs-low-fidelity}, these patches differ in texture.
Given pairs
$\boldsymbol{y}, \boldsymbol{x} \sim \widehat{\pi}_{y, x} (\boldsymbol{y}, \boldsymbol{x})$,
we pretrain our network by minimizing Equation (\ref{hint3-obj}).
Figures~\ref{denoising-true} and~\ref{denoising-obs} contain a pair not
used during pretraining. Estimates for the conditional mean and standard
deviation obtained by drawing $1000$ samples from the pretrained
conditional NF for the noisy indirect measurement
(Figure~\ref{denoising-obs}) are included in Figures~\ref{denoising-cm}
and~\ref{denoising-std}. Both estimates exhibit the expected behavior
because the examples in Figure~\ref{denoising-true}
and~\ref{denoising-obs} are within the distribution. As anticipated,
this observation no longer holds if we apply this pretrained network to
indirect data depicted in Figure~\ref{denoising-WS-obs}, which is
sampled from the deeper part. However, these results are significantly
improved when the pretrained network is fine-tuned by minimizing
Equation (\ref{hint2-obj-finetune}). After fine tuning, the fine details
in the image are recovered (compare
Figures~\ref{denoising-cm-pretrained} and~\ref{denoising-cm-ws}). This
improvement is confirmed by the relative errors plotted in
Figures~\ref{denoising-cm-err} and~\ref{denoising-cm-ws-err}, as well as
by the reduced standard deviation (compare
Figures~\ref{denoising-std-ws-pretrained} and~\ref{denoising-std-ws}).

\section{Conclusions}\label{conclusions}

Inverse problems in fields such as seismology are challenging for
several reasons. The forward operators are complex and expensive to
evaluate numerically while the Earth is highly heterogeneous. To handle
this situation and to quantify uncertainty, we propose a preconditioned
scheme for training normalizing flows for Bayesian inference. The
proposed scheme is designed to take full advantage of having access to
training pairs drawn from a joint distribution, which for the reasons
stated above is close but not equal to the actual joint distribution. We
use these samples to train a normalizing flow via likelihood
maximization leveraging the normalizing property. We use this pretrained
low-fidelity estimate for the posterior as a prior and preconditioner
for the actual variational inference on the observed data, which
minimizes the Kullback-Leibler divergence between the predicted and the
desired posterior density. By means of a series of examples, we
demonstrate that our preconditioned scheme leads to considerable
speed-ups compared to training a normalizing flow from scratch.

\bibliography{siahkoohi2020ABIfab}

\appendix

\clearpage

\section{Mathematical derivations}\label{mathematical-derivations}

Let $f : \mathcal{Z} \to \mathcal{X}$ be a bijective transformation that
maps a random variable $\boldsymbol{z} \sim \pi_{z} (\boldsymbol{z})$ to
$\boldsymbol{x} \sim \pi_{x} (\boldsymbol{x})$. We can write the change
of variable formula \citep{villani2009optimal} that relates probability
density functions $\pi_{z}$ and $\pi_{x}$ in the following manner:
\begin{equation}
\pi_x (\boldsymbol{x}) = \pi_z (\boldsymbol{z})\,
    \Big | \det \nabla_{x} f^{-1} (\boldsymbol{x}) \Big |, \quad f (\boldsymbol{z}) = \boldsymbol{x}, \quad \boldsymbol{x} \in \mathcal{X}.
\label{change-of-variable}
\end{equation}
 This relation serves as the basis for the objective functions used
throughout this paper.

\subsection{\texorpdfstring{Derivation of Equation
(\ref{hint2-obj})}{Derivation of Equation ()}}\label{derivation-of-equation-hint2-obj}

In Equation~\ref{hint2-obj}, we train a bijective transformation,
denoted by $T_{\theta} : \mathcal{Z}_x \to \mathcal{X}$, that maps the
latent distribution $\pi_{z_x}(\boldsymbol{z}_x)$ to the high-fidelity
posterior density
$\pi_{\text{post}} (\boldsymbol{x} \mid \boldsymbol{y} )$. We optimize
the parameters of $T_{\theta}$ by minimizing the KL divergence between
the push-forward density \citep{bogachev2006measure}, denoted by
$\pi_{\theta} (\, \cdot \mid \boldsymbol{y}) := T_{\sharp}\, \pi_{z_x}$,
and the posterior density:
\begin{equation}
\begin{aligned}
& \mathop{\rm arg\,min}_{\theta}\,
    \mathbb{D}_{\text{KL}} \left (\pi_{\theta} (\, \cdot \mid
    \boldsymbol{y}) \mid\mid \pi_{\text{post}} (\, \cdot \mid
    \boldsymbol{y}) \right ) \\
= & \mathop{\rm arg\,min}_{\theta}\, \mathbb{E}_{\boldsymbol{x}
    \sim \pi_{\theta} (\boldsymbol{x} \mid \boldsymbol{y})}
    \Big [ - \log \pi_{\text{post}} (\boldsymbol{x} \mid
    \boldsymbol{y} ) + \log \pi_{\theta} (\boldsymbol{x} \mid
    \boldsymbol{y}) \Big ].
\end{aligned}
\label{hint2-obj-derivation}
\end{equation}
 In the above expression, we can rewrite the expectation with respect to
$\pi_{\theta} (\boldsymbol{x} \mid \boldsymbol{y})$ as the expectation
with respect to the latent distribution, followed by a mapping via
$T_{\theta}$---i.e.,
\begin{equation}
\mathop{\rm arg\,min}_{\theta}\, \mathbb{E}_{\boldsymbol{z}_x
    \sim \pi_{z_x} (\boldsymbol{z})}
    \Big [ - \log \pi_{\text{post}} (T_{\theta} (\boldsymbol{z}_x) \mid
    \boldsymbol{y} )
    + \log \pi_{\theta} (T_{\theta} (\boldsymbol{z}_x) \mid \boldsymbol{y}) \Big ].
\label{change-expectation}
\end{equation}
 The last term in the expectation above can be further simplified via
the change of variable formula in Equation (\ref{change-of-variable}).
If $\boldsymbol{x} = T_{\theta} (\boldsymbol{z}_x)$, then:
\begin{equation}
\pi_{\theta} (\boldsymbol{x} \mid \boldsymbol{y}) = \pi_{z_x}
    (\boldsymbol{z}_x)\,
    \Big | \det \nabla_{x} T_{\theta} ^{-1} (\boldsymbol{x}) \Big |
    = \pi_{z_x} (\boldsymbol{z}_x)\,
    \Big | \det \nabla_{z_x} T_{\theta} (\boldsymbol{z}_x) \Big |^{-1}.
\label{change-of-variable-hint2}
\end{equation}
 The last equality in Equation (\ref{change-of-variable-hint2}) holds
due to the invertibility of $T_{\theta}$ and the differentiability of
its inverse (inverse function theorem). By combining Equations
(\ref{change-expectation}) and (\ref{change-of-variable-hint2}), we
arrive at the following objective function for training $T_{\theta}$:

\begin{equation}
\mathop{\rm arg\,min}_{\theta}\,
    \mathbb{E}_{\boldsymbol{z}_x
    \sim \pi_{z_x} (\boldsymbol{z})}
    \Big [ - \log \pi_{\text{post}} (T_{\theta} (\boldsymbol{z}_x) \mid
    \boldsymbol{y} ) + \log \pi_{z_x} (\boldsymbol{z}_x)
    -\log  \Big | \det
    \nabla_{z_x} T_{\theta} (\boldsymbol{z}_x) \Big | \Big ].
\label{final-equ-hint2}
\end{equation}

Finally, by ignoring the $\log \pi_{z_x} (\boldsymbol{z}_x)$ term, which
is constant with respect to $\boldsymbol{\theta}$, using Bayes' rule,
and inserting our data likelihood model from Equation (\ref{fwd-op}), we
derive Equation (\ref{hint2-obj}):

\begin{equation}
\min_{\theta}\, \mathbb{E}\,_{\boldsymbol{z}_x \sim
    \pi_{z_x}(\boldsymbol{z}_x)}
    \bigg [ \frac{1}{2\sigma^2}  \left \| F \big(T_{\theta}
    (\boldsymbol{z}_x) \big) -
    \boldsymbol{y} \right \|_2^2  -\log \pi_{\text{prior}}
    \big (T_{\theta} (\boldsymbol{z}_x) \big ) -\log  \Big | \det
    \nabla_{z_x} T_{\theta} (\boldsymbol{z}_x) \Big | \bigg ].
\label{hint2-obj-derived}
\end{equation}

Next, based on this equation, we derive the objective function used in
the pretraining phase.

\subsection{\texorpdfstring{Derivation of Equation
(\ref{hint3-obj})}{Derivation of Equation ()}}\label{derivation-of-equation-hint3-obj}

The derivation of objective in Equation (\ref{hint3-obj}) follows
directly from the change of variable formula in
Equation~\ref{change-of-variable}, applied to a bijective map,
$G^{-1}_{\phi} : \mathcal{Z}_y \times \mathcal{Z}_x \to \mathcal{Y} \times \mathcal{X}$,
where $\mathcal{Z}_y$ and $\mathcal{Z}_y$ are Gaussian latent spaces.
That is to say:
\begin{equation}
\begin{split}
\widehat{\pi}_{y, x} (\boldsymbol{y}, \boldsymbol{x}) =
    \pi_{z_y, z_x} (\boldsymbol{z}_y, \boldsymbol{z}_x)\,
    \Big | \det \nabla_{y, x}  G_{\phi} (\boldsymbol{y}, \boldsymbol{x})
    \Big |, \quad G_{\phi}(\boldsymbol{y}, \boldsymbol{x})
    =\begin{bmatrix}  \boldsymbol{z}_y \\\boldsymbol{z}_x \end{bmatrix}.
\end{split}
\label{change-of-variable-hint3}
\end{equation}
 Given (low-fidelity) training pairs,
$\boldsymbol{y}, \boldsymbol{x}  \sim \widehat{\pi}_{y, x} (\boldsymbol{y}, \boldsymbol{x})$,
the maximum likelihood estimate of $\boldsymbol{\phi}$ is obtained via
the following objective:
\begin{equation}
\begin{aligned}
& \mathop{\rm arg\,max}_{\phi}\, \mathbb{E}\,_{\boldsymbol{y},
    \boldsymbol{x} \sim \widehat{\pi}_{y, x} (\boldsymbol{y}, \boldsymbol{x})}\, \left [ \log \widehat{\pi}_{y, x} (\boldsymbol{y}, \boldsymbol{x}) \right ] \\
& = \mathop{\rm arg\,min}_{\phi}\, \mathbb{E}\,_{\boldsymbol{y}, \boldsymbol{x}
    \sim \widehat{\pi}_{y, x} (\boldsymbol{y}, \boldsymbol{x})}\,
    \left [ - \log \pi_{z_y, z_x} (\boldsymbol{z}_y, \boldsymbol{z}_x)
    - \log \Big|\det \nabla_{y, x}\, G_{\phi} (\boldsymbol{y}, \boldsymbol{x}) \Big | \right ] \\
& = \mathop{\rm arg\,min}_{\phi}\, \mathbb{E}\,_{\boldsymbol{y}, \boldsymbol{x}
    \sim \widehat{\pi}_{y, x} (\boldsymbol{y}, \boldsymbol{x})}\,
    \left [ \frac{1}{2} \left \| G_{\phi} (\boldsymbol{y}, \boldsymbol{x}) \right \|^2 - \log \Big|\det \nabla_{y, x}\, G_{\phi} (\boldsymbol{y}, \boldsymbol{x}) \Big | \right ],
\end{aligned}
\label{mle-hint3}
\end{equation}
 that is the objective function in Equation (\ref{hint3-obj}). The NF
trained via the objective function, given samples from the latent
distribution, draws samples from the low-fidelity joint distribution,
$\widehat{\pi}_{y, x}$.

By construction, $G_{\phi}$ is a block-triangular map---i.e.,
\begin{equation}
\begin{split}
\quad G_{\phi}(\boldsymbol{y}, \boldsymbol{x})
    =\begin{bmatrix}  G_{\phi_y} (\boldsymbol{y}) \\G_{\phi_x}
    (\boldsymbol{y}, \boldsymbol{x}) \end{bmatrix}, \
    \boldsymbol{\phi} = \left \{ \boldsymbol{\phi}_y ,
    \boldsymbol{\phi}_x \right \}.
\end{split}
\label{KT-map}
\end{equation}
 \citet{kruse2019hint} showed that after solving the optimization
problem in Equation (\ref{hint3-obj}), $G_{\phi}$ approximates the
well-known triangular Knothe-Rosenblat map
\citep{santambrogio2015optimal}. As shown in
\citet{marzouk2016sampling}, the triangular structure and $G_{\phi}$'s
invertibility yields the following property,
\begin{equation}
\left (G_{\phi_x}^{-1} (G_{\phi_y} (\boldsymbol{y}), \cdot \,) \right)_{\sharp}\, \pi_{z_x} = \widehat{\pi}_{\text{post}} (\, \cdot \mid \boldsymbol{y}),
\label{sampling-KT}
\end{equation}
 where $\widehat{\pi}_{\text{post}}$ denotes the low-fidelity posterior
probability density function. The expression above means we can get
access to low-fidelity posterior distribution samples by simply
evaluating
$G_{\phi_x}^{-1} (G_{\phi_y} (\boldsymbol{y}), \boldsymbol{z}_x)$ for
$\boldsymbol{z}_x \sim \pi_{z_x}(\boldsymbol{z}_x)$ for a given observed
data $\boldsymbol{y}$.

\section{Training details and network
architectures}\label{training-details-and-network-architectures}

For our network architecture, we adapt the recursive coupling blocks
proposed by \citet{kruse2019hint}, which use invertible coupling layers
from \citet{dinh2016density} in a hierarchical way. In other words, we
recursively divide the incoming state variables and apply an affine
coupling layer. The final architecture is obtained by composing several
of these hierarchical coupling blocks. The hierarchical structure leads
to dense triangular Jacobian, which is essential in representation power
of NFs \citep{kruse2019hint}.

For all examples in this paper, we use the hierarchical coupling blocks
as described in \citet{kruse2019hint}. The affine coupling layers within
each hierarchal block contain a residual block as described in
\citet{he2016deep}. Each residual block has the following dimensions:
$64$ input, $128$ hidden, and $64$ output channels, except for the first
and last coupling layer where we have $4$ input and output channels,
respectively. We use the Wavelet transform and its transpose before
feeding seismic images into the network and after the last layer of the
NFs.

Below, we describe the network architectures and training details
regarding the two numerical experiments described in the paper.
Throughout the experiments, we use the Adam optimization algorithm
\citep{kingma2015adam}.

\subsection{2D toy example}\label{d-toy-example-1}

We use 8 hierarchal coupling blocks, as described above for both
$G_{\phi_x}$ and $G_{\phi_y}$ (Equation (\ref{hint3-obj})). As a result,
due to our proposed method in Equation (\ref{low-fidelity-NF}), we
choose the same architecture for $T_{\phi_x}$ (Equation
(\ref{hint2-obj})).

For pretraining $G_{\theta}$ according to Equation (\ref{hint3-obj}), we
use $5000$ low-fidelity joint training pairs,
$\boldsymbol{y}, \boldsymbol{x} \sim \widehat{\pi}_{y, x} (\boldsymbol{y}, \boldsymbol{x})$.
We minimize Equation (\ref{hint3-obj}) for $25$ epochs with a batch size
of $64$ and a (starting) learning rate of $0.001$. We decrease the
learning rate each epoch by a factor of $0.9$.

For the preconditioned step---i.e., solving
Equation~\ref{hint2-obj-finetune}, we use $1000$ latent training
samples. We train for $5$ epochs with a batch size of $64$ and a
learning rate $0.001$. Finally, as a comparison, we solve the objective
in Equation~\ref{hint2-obj-finetune} for a randomly initialized NF with
the same $1000$ latent training samples for $25$ epochs. We decrease the
learning rate each epoch by $0.9$.

\subsection{Seismic compressed
sensing}\label{seismic-compressed-sensing}

We use 12 hierarchal coupling blocks, as described above for both
$G_{\phi_x}$, $G_{\phi_y}$, and we use the same architecture for
$T_{\phi_x}$ as $G_{\phi_x}$.

For pretraining $G_{\theta}$ according to Equation (\ref{hint3-obj}), we
use $5282$ low-fidelity joint training pairs,
$\boldsymbol{y}, \boldsymbol{x} \sim \widehat{\pi}_{y, x} (\boldsymbol{y}, \boldsymbol{x})$.
We minimize Equation (\ref{hint3-obj}) for $50$ epochs with a batch size
of $4$ and a starting learning rate of $0.001$. Once again, we decrease
the learning rate each epoch by $0.9$.

For the preconditioned step---i.e., solving
Equation~\ref{hint2-obj-finetune}, we use $1000$ latent training
samples. We train for $10$ epochs with a batch size $4$ and a learning
rate of $0.0001$, where we decay the step by $0.9$ in every $5$th epoch.

\section{2D toy example---more results}\label{d-toy-examplemore-results}

\begin{figure}
\centering
\captionsetup[subfigure]{labelformat=empty}
\subfloat[]{\includegraphics[width=0.250\hsize]{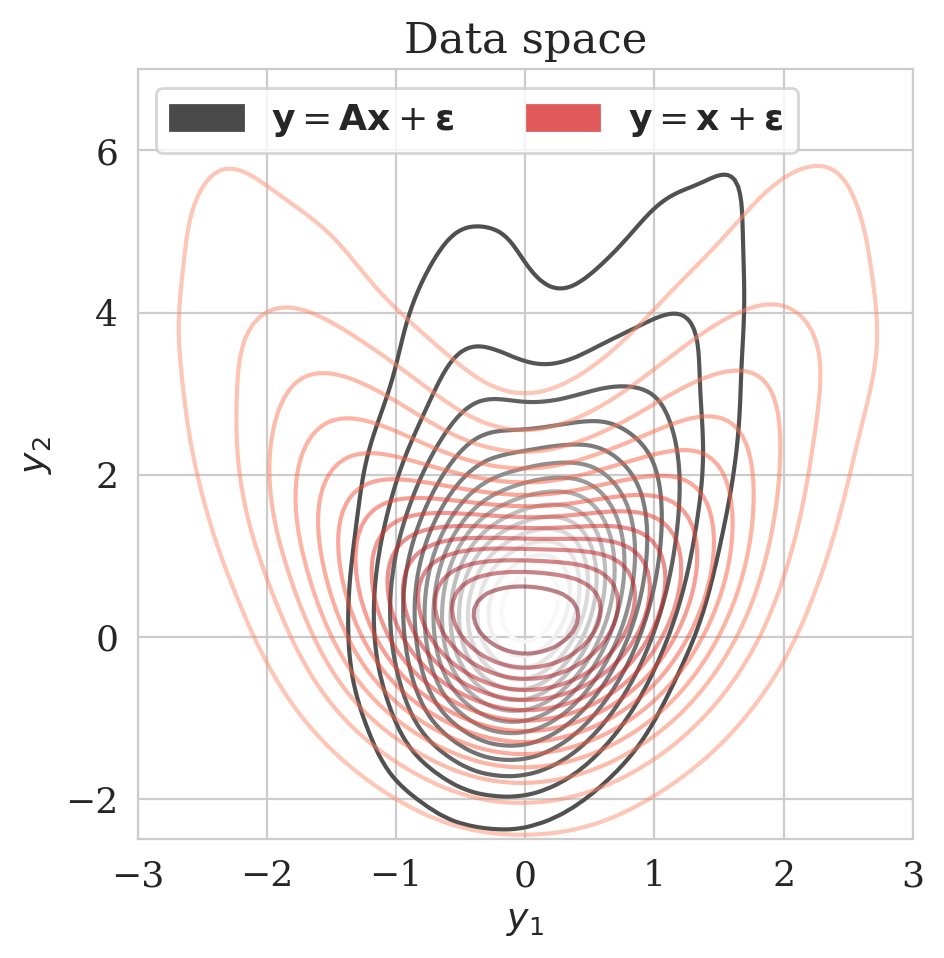}}
\subfloat[]{\includegraphics[width=0.250\hsize]{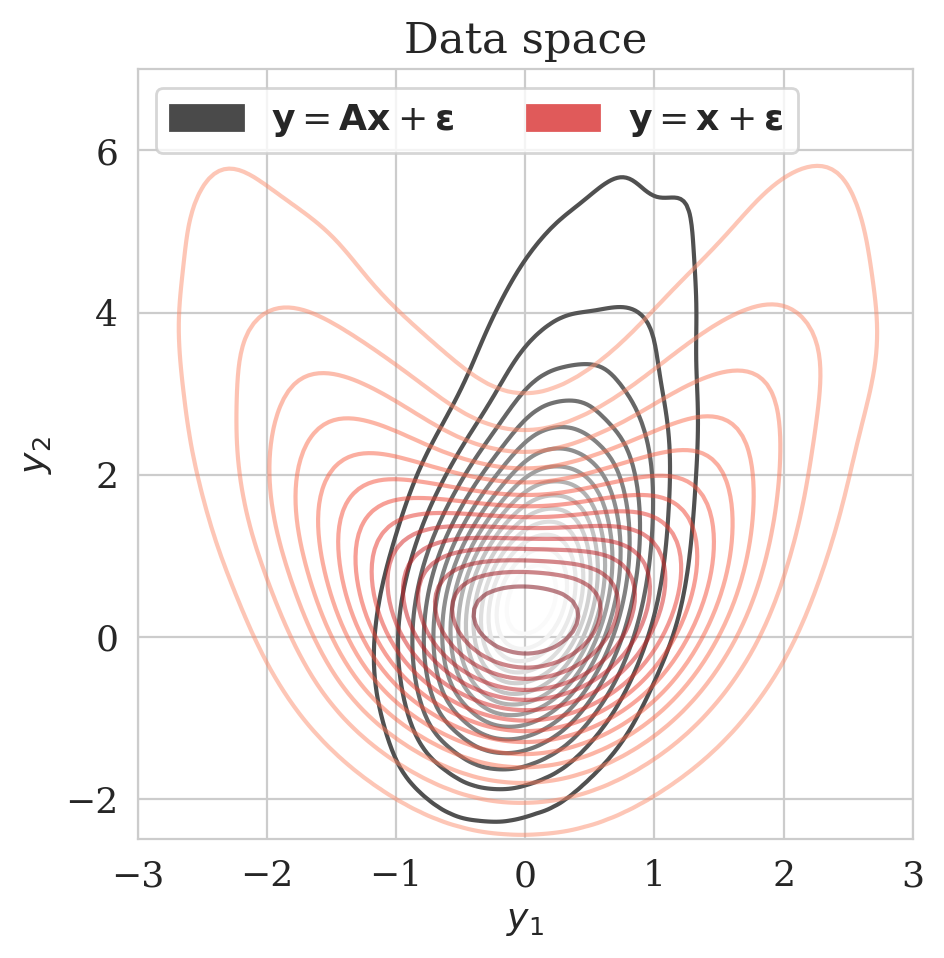}}
\subfloat[]{\includegraphics[width=0.250\hsize]{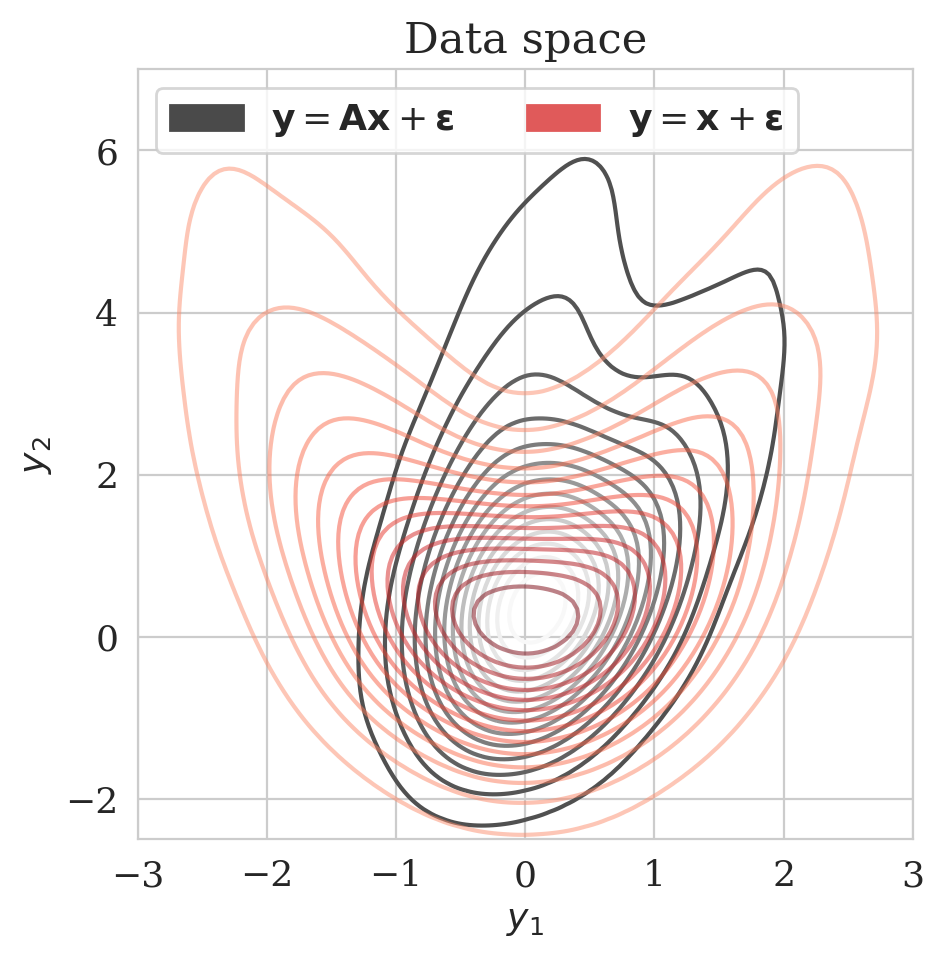}}
\\
\subfloat[]{\includegraphics[width=0.250\hsize]{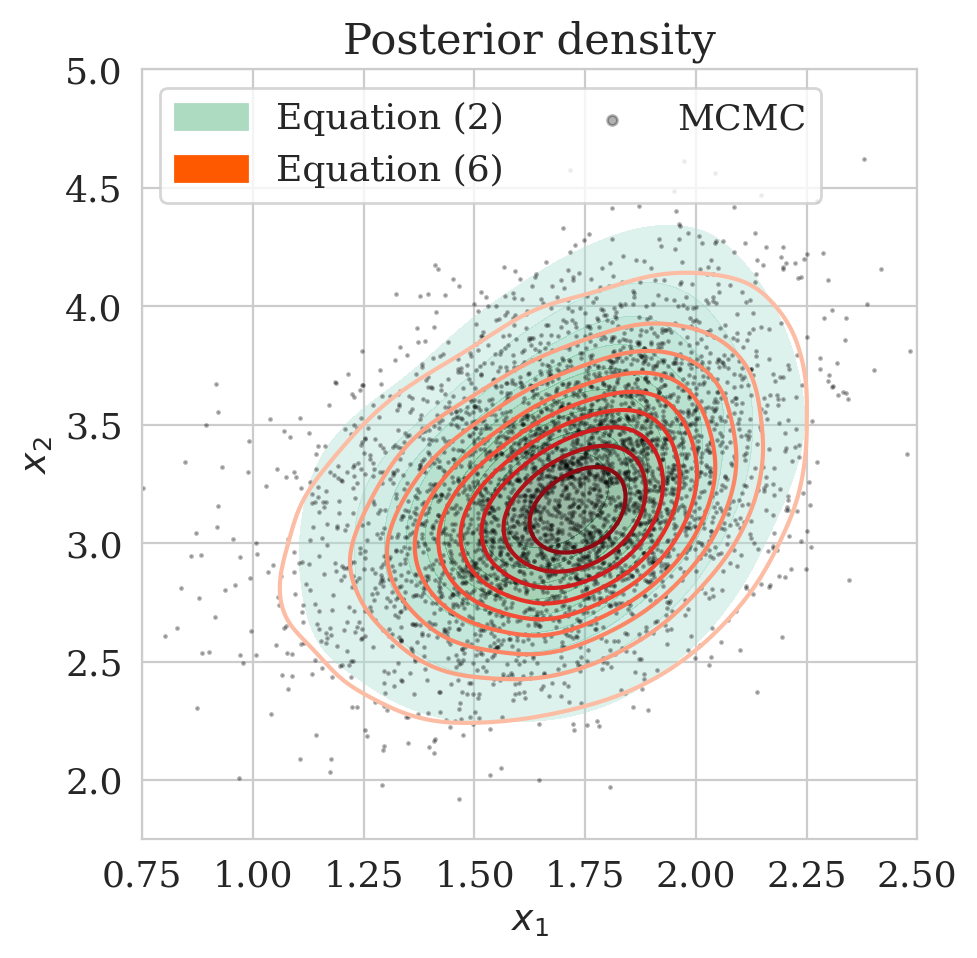}}
\subfloat[]{\includegraphics[width=0.250\hsize]{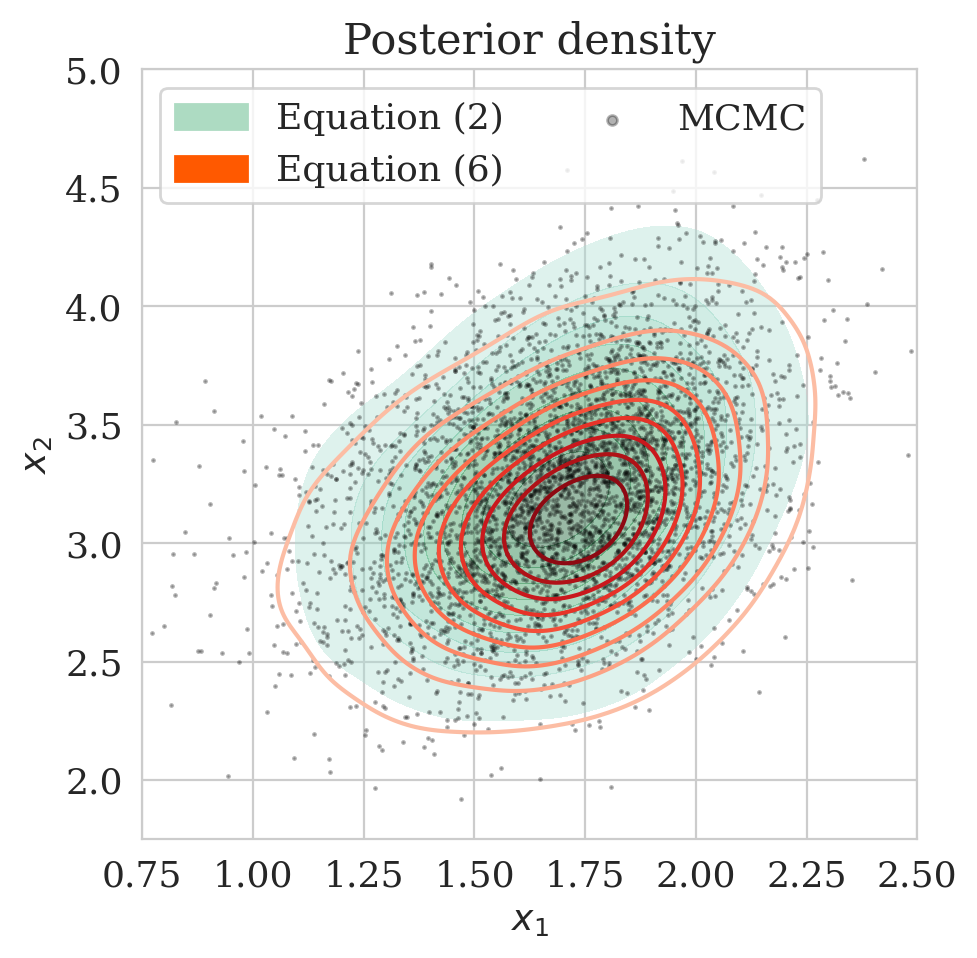}}
\subfloat[]{\includegraphics[width=0.250\hsize]{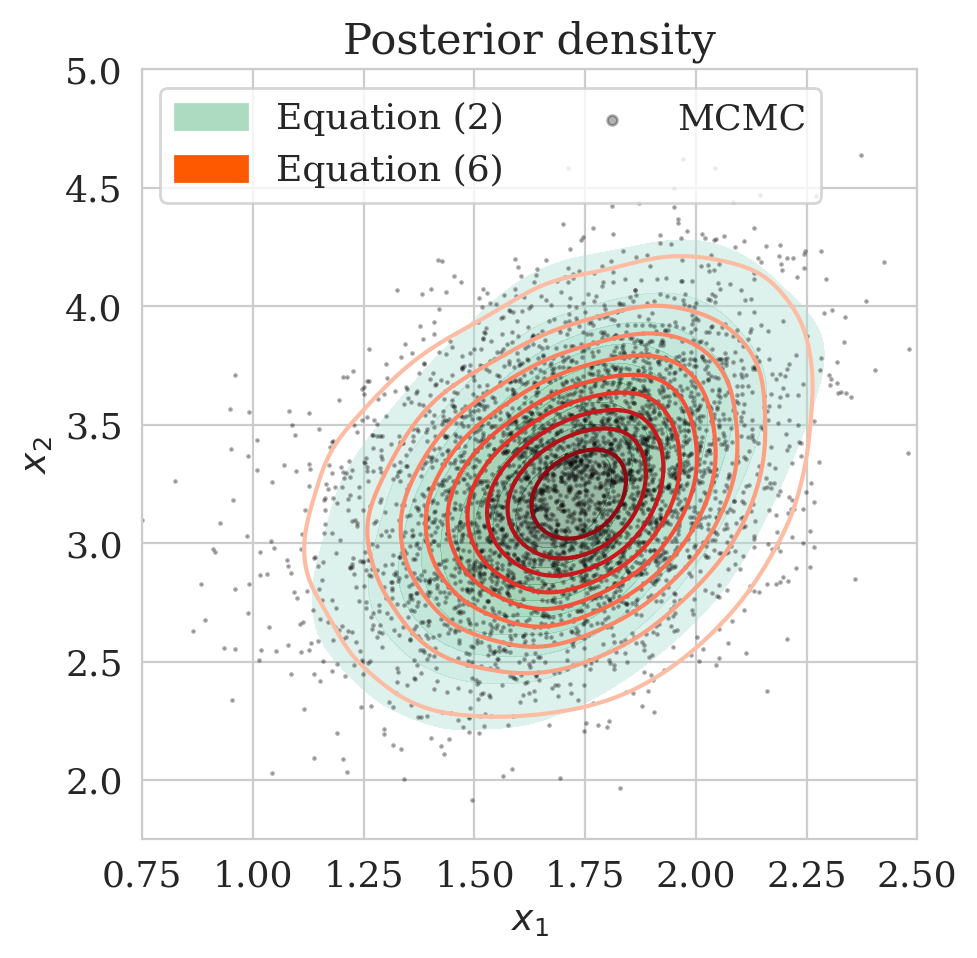}}
\\
\subfloat[]{\includegraphics[width=0.250\hsize]{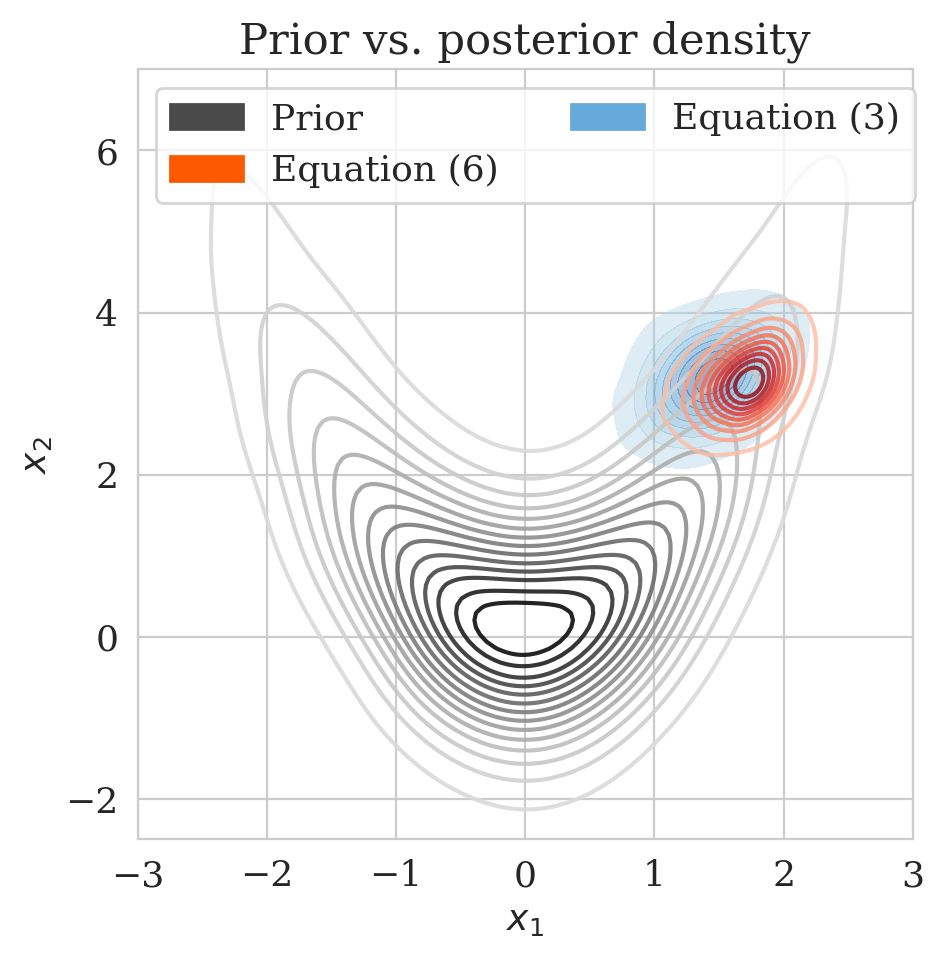}}
\subfloat[]{\includegraphics[width=0.250\hsize]{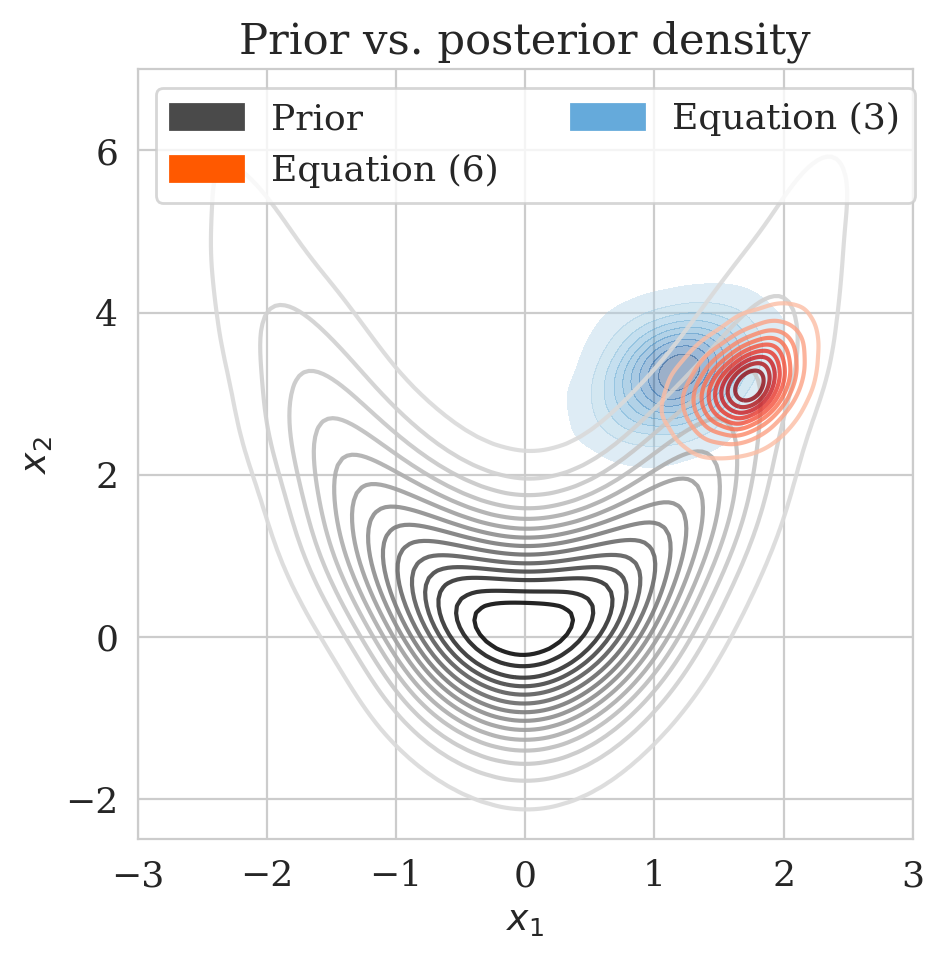}}
\subfloat[]{\includegraphics[width=0.250\hsize]{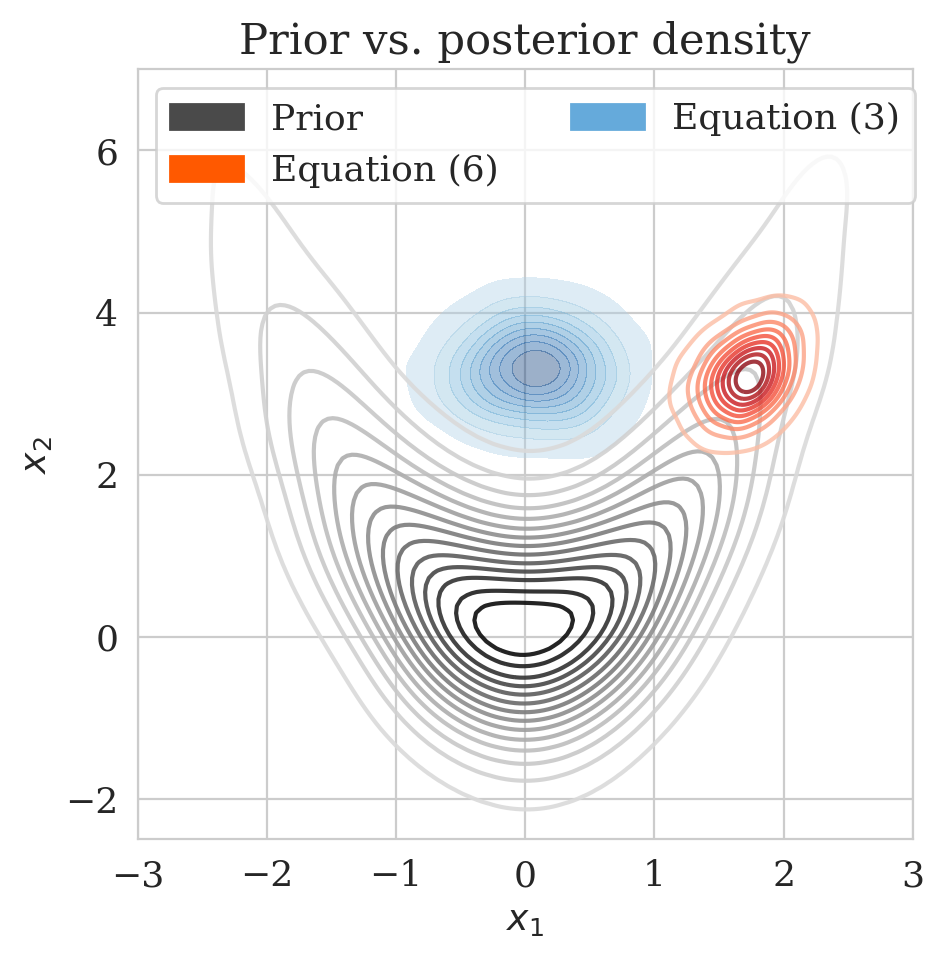}}
\\
\subfloat[]{\includegraphics[width=0.250\hsize]{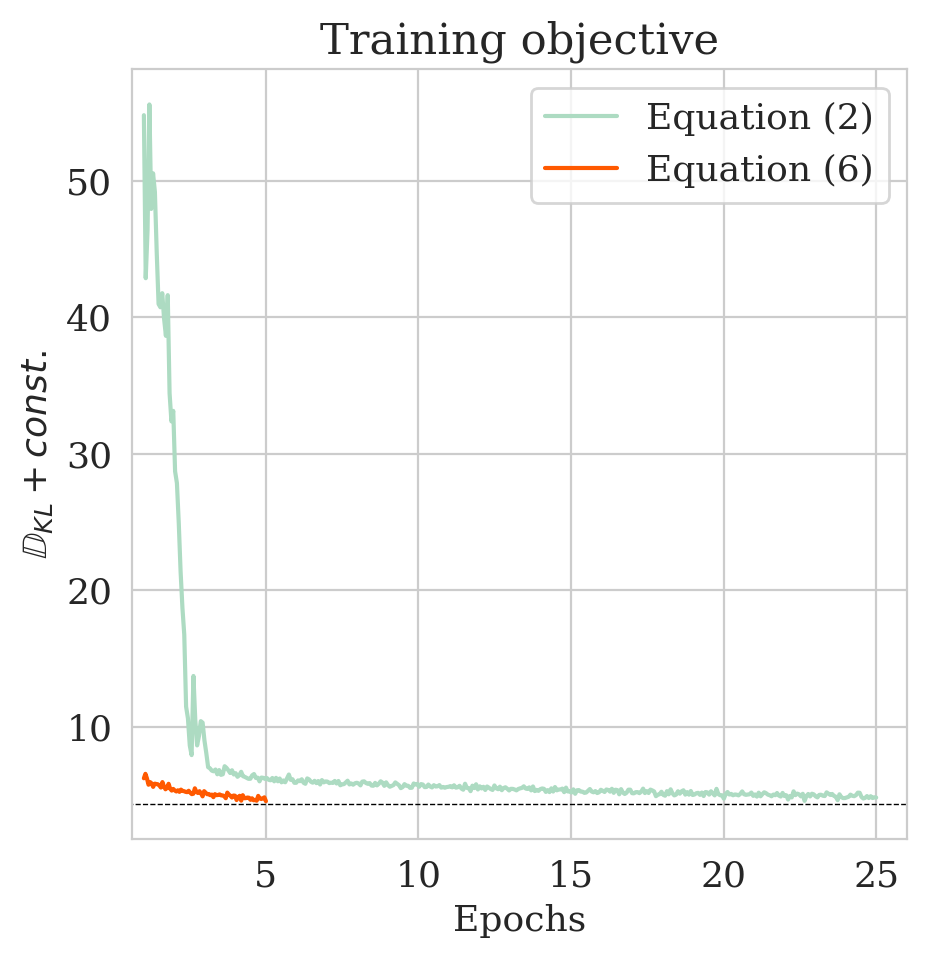}}
\subfloat[]{\includegraphics[width=0.250\hsize]{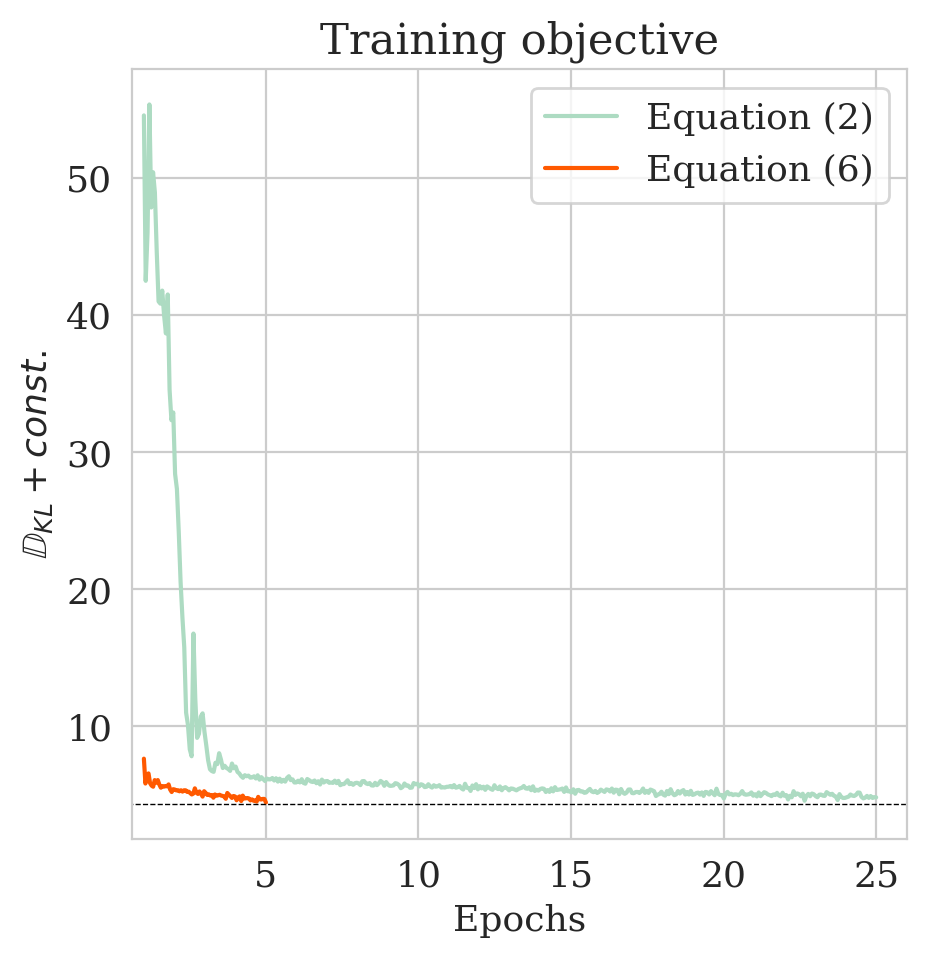}}
\subfloat[]{\includegraphics[width=0.240\hsize]{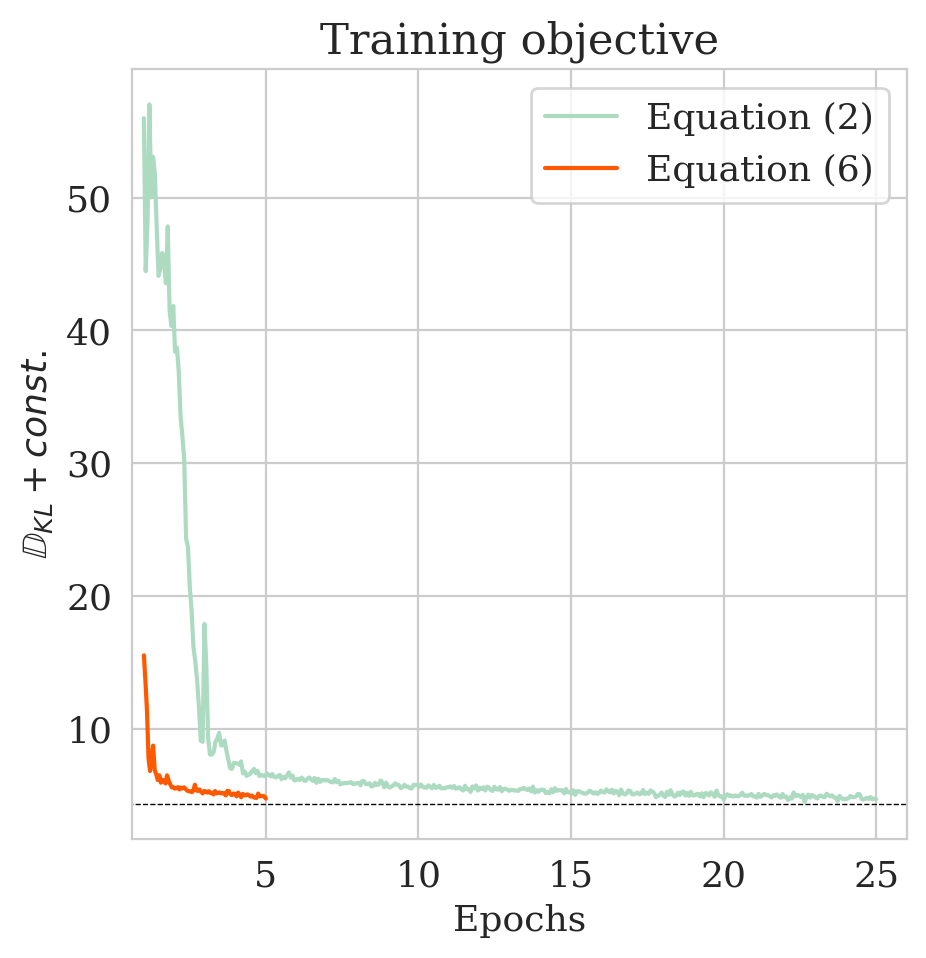}}
\caption{2D toy example with decreasing values of $\gamma = 2, 1, 0$
from first to last column, respectively. First row: Low- and
high-fidelity data. Second row: approximated posterior densities via
MCMC (dark circles), and objectives in Equations (\ref{hint2-obj}) and
(\ref{hint2-obj-finetune}). Third row: overlaid prior density with
predicted posterior densities via objectives in Equations
(\ref{hint3-obj}) and (\ref{hint2-obj-finetune}). Last row: training
objective values during training via Equations (\ref{hint2-obj}) and
(\ref{hint2-obj-finetune}).}\label{gammaexps}
\end{figure}

Here we show the effect $\gamma$ on our proposed method in the 2D toy
experiment. By choosing smaller values for $\gamma$, we make
$\bar{\boldsymbol{A}} \mathbin{/} \rho (\bar{\boldsymbol{A}})$ with
$\bar{\boldsymbol{A}} = \boldsymbol{\Gamma} + \gamma \boldsymbol{I}$
less close to the identity matrix, hence enhancing the discrepancy
between the low- and high-fidelity posterior. The first row of
Figure~\ref{gammaexps} shows the low- (purple) and high-fidelity (red)
data densities for decreasing values of $\gamma$ from $2$ down to $0$.
The second row depicts the predicted posterior densities via the
preconditioned scheme (orange contours) and the low-fidelity posterior
in green along with MCMC samples (dark circles). The third row compares
the preconditioned posterior densities to samples obtained via the
low-fidelity pretrained NF---i.e., Equation (\ref{hint3-obj}). Finally,
the last row shows the objective function values during training with
(orange) and without (green) preconditioning.

We observe that by decreasing $\gamma$ from $2$ to $0$, the low-fidelity
posterior approximations become worse. As a result, the objective
function for the preconditioned approach (orange) at the beginning start
from a higher value, indicating more mismatch between low- and
high-fidelity posterior densities. Finally, our preconditioning method
consistently improves upon low-fidelity posterior by training for $5$
epochs.

\section{Seismic compressed sensing---more
results}\label{seismic-compressed-sensingmore-results}

\begin{figure}
\centering
\captionsetup[subfigure]{labelformat=empty}
\subfloat[]{\includegraphics[width=0.250\hsize]{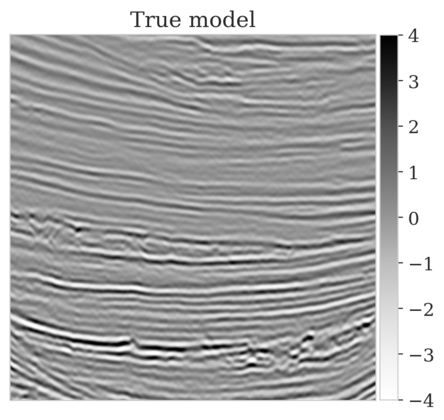}}
\subfloat[]{\includegraphics[width=0.250\hsize]{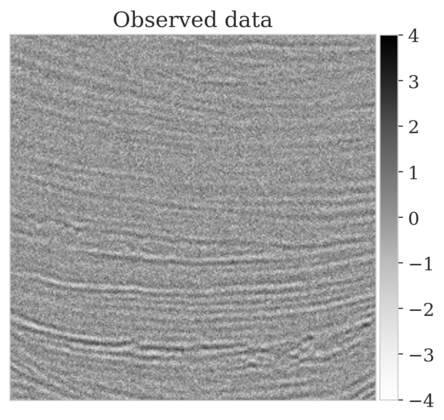}}
\subfloat[]{\includegraphics[width=0.250\hsize]{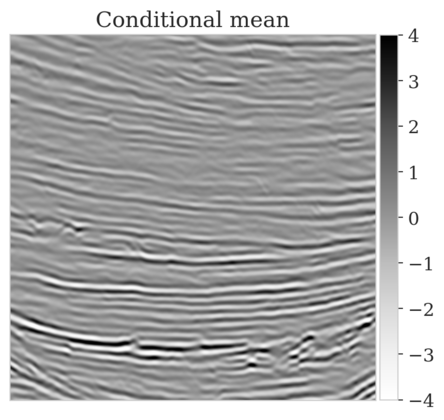}}
\subfloat[]{\includegraphics[width=0.250\hsize]{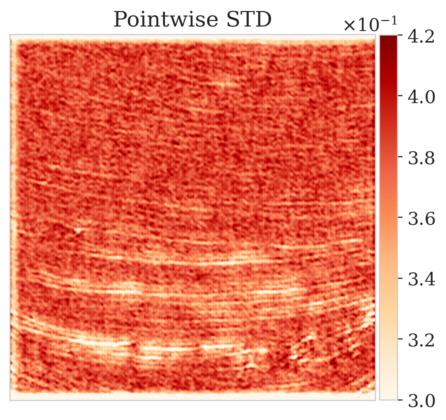}}
\\
\subfloat[]{\includegraphics[width=0.250\hsize]{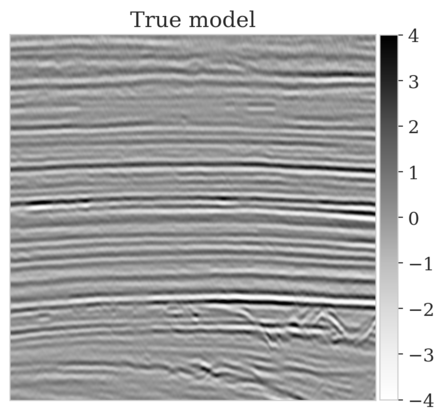}}
\subfloat[]{\includegraphics[width=0.250\hsize]{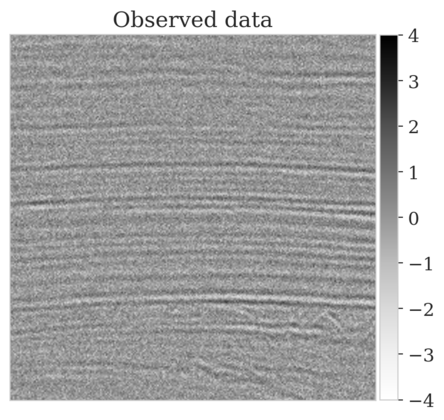}}
\subfloat[]{\includegraphics[width=0.250\hsize]{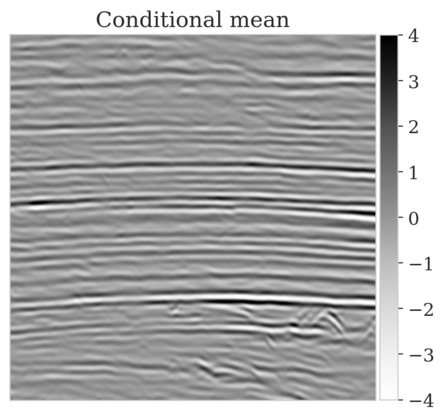}}
\subfloat[]{\includegraphics[width=0.250\hsize]{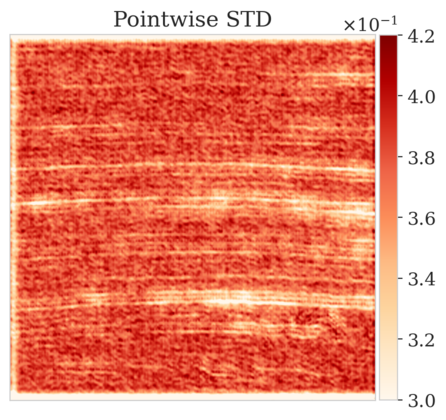}}
\\
\subfloat[]{\includegraphics[width=0.250\hsize]{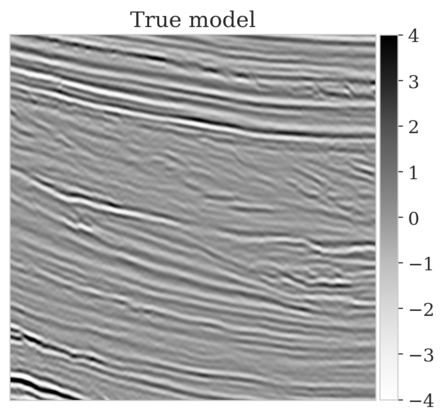}}
\subfloat[]{\includegraphics[width=0.250\hsize]{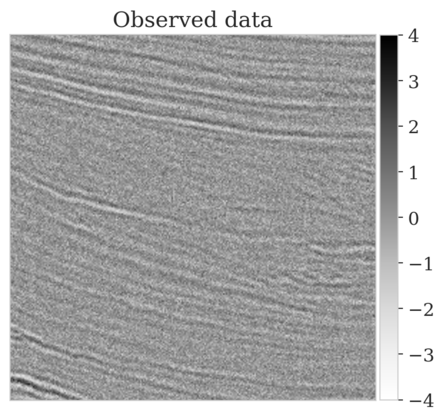}}
\subfloat[]{\includegraphics[width=0.250\hsize]{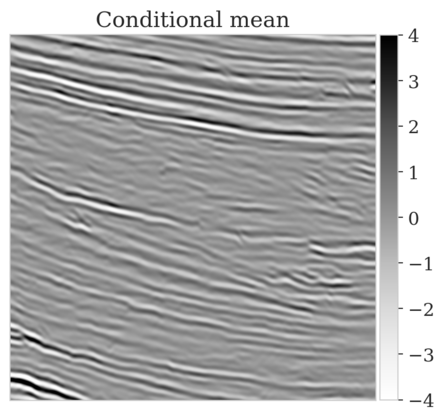}}
\subfloat[]{\includegraphics[width=0.250\hsize]{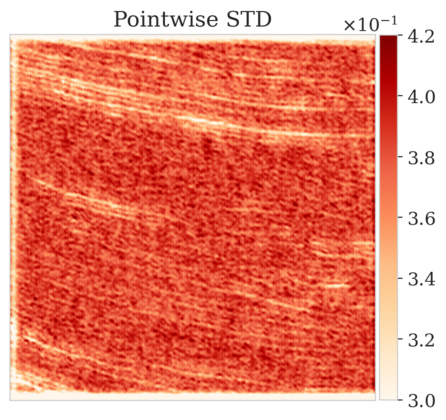}}
\\
\subfloat[]{\includegraphics[width=0.250\hsize]{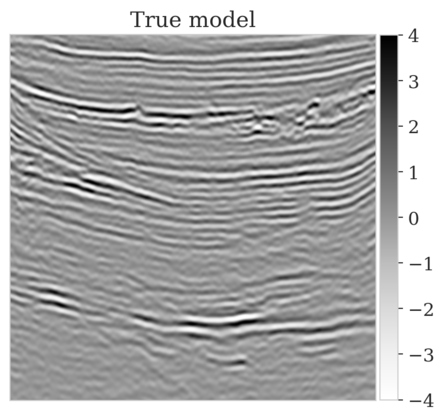}}
\subfloat[]{\includegraphics[width=0.250\hsize]{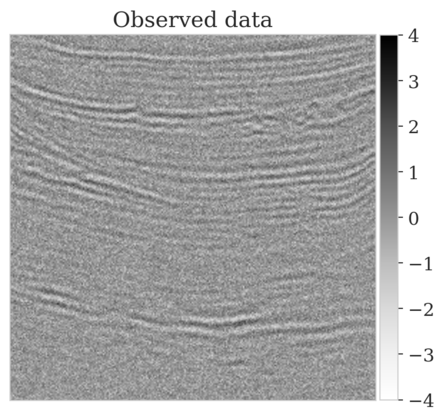}}
\subfloat[]{\includegraphics[width=0.250\hsize]{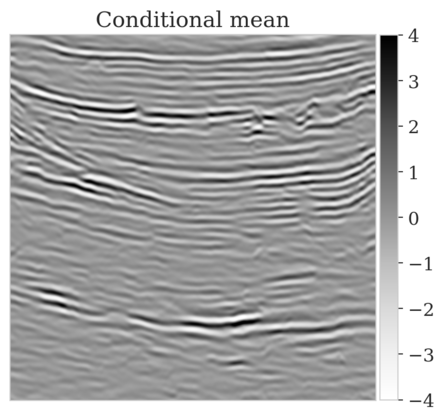}}
\subfloat[]{\includegraphics[width=0.250\hsize]{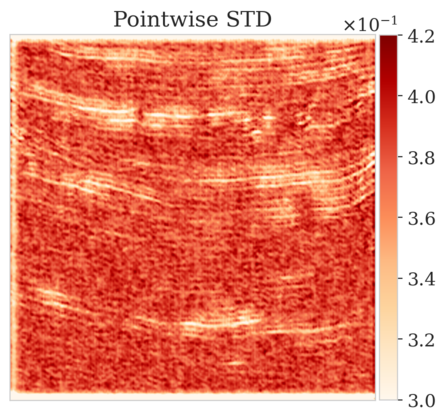}}
\caption{Seismic compressed sensing for four different low-fidelity
images. First column: true seismic images. Second column: low-fidelity
observed data. Third and last columns: conditional mean and pointwise
STD estimates obtained by drawing $1000$ samples from the pretrained
conditional NF.}\label{denoising-appendix-more-seismic-images}
\end{figure}

Here we show more examples to verify the pretraining phase obtained via
solving the objective in Equation (\ref{hint2-obj}). Each row in
Figure~\ref{denoising-appendix-more-seismic-images} corresponds to a
different testing image. The first column shows the different true
seismic images used to create low-fidelity compressive sensing data,
depicted in the second column. The third and last columns correspond to
the conditional mean and pointwise STD estimates, respectively. Clearly,
the pretrained network is able to successfully recover the true image,
and consistently indicates more uncertainty in areas with low-amplitude
events.

\typeout{get arXiv to do 4 passes: Label(s) may have changed. Rerun}
\end{document}